%% file: main.tex
\documentclass[11pt]{article}

\usepackage{algorithm}
\usepackage{algorithmic}
\usepackage{placeins}
\usepackage{float}
\usepackage{threeparttable}
\usepackage{tabularx}
\usepackage{needspace}
\usepackage{agenticlearning-xelatex}


\begin{document}

\shorttitle{Seeing the City or Recognizing the Place?}
\shortauthor{Kaizhen Tan}

\title{Seeing the City or Recognizing the Place?\\[3pt]
{\large What Street-View Imagery Adds Beyond Existing Urban Data\\[1pt]in VLM-Based Urban Sensing}}
\author{Kaizhen Tan\\[2pt]
{\normalsize New York University}}
\date{}
\maketitle

\input{sec/0_abstract}
\input{sec/1_intro}
\input{sec/2_related}
\input{sec/3_taxonomy}
\input{sec/4_protocol}
\input{sec/5_study1}
\input{sec/6_study2}
\input{sec/7_study3}
\input{sec/8_discussion}

\section*{Data Availability}
The companion \texttt{reproduction\_archive.zip} contains analysis scripts,
raw model responses, sample identifiers, figure code, and literature-review
records, and it documents the models, prompts, run dates, and sampling
choices used in the experiments. Third-party data remain available from the
original sources under their respective terms; the archive includes source
identifiers and retrieval scripts that reacquire them from Overpass, Project
Sidewalk, Seattle ArcGIS, Zenodo, and CityLens.

\section*{Declaration of Generative AI Use}
The author used OpenAI Codex and Claude Code (Anthropic) for language
editing and revision suggestions, including literature suggestions that the
author checked against the original sources. The author reviewed the content
and takes full responsibility for the manuscript.

\bibliographystyle{apalike}
\bibliography{main}

\clearpage
\setcounter{table}{0}
\setcounter{figure}{0}
\setcounter{equation}{0}
\renewcommand{\thetable}{S\arabic{table}}
\renewcommand{\thefigure}{S\arabic{figure}}
\renewcommand{\theequation}{S\arabic{equation}}
\crefname{appendix}{Supplementary Section}{Supplementary Sections}
\Crefname{appendix}{Supplementary Section}{Supplementary Sections}
\input{sec/9_app}

\end{document}

%% file: sec/0_abstract.tex
\begin{abstract}
Street-view imagery is increasingly analysed with vision-language models (VLMs) to infer urban attributes, but predictive accuracy alone does not show how much a photograph contributes beyond data already available for the same place. Using three VLMs, we compare image-based predictions with existing urban data for seven attributes drawn from five public resources. Each urban unit is evaluated with imagery, with location or text context, and against non-image predictions from nearby observations or public records. We also replace images and add conflicting records to test which source the predictions follow. Nearby observations or public records matched or exceeded image-only predictions for road damage, curb ramps, population, and house price. Images were more informative for building type, building function, and the floor count of low-rise buildings. For floor count, the advantage of images over nearby OpenStreetMap labels increased by 5.7 percentage points per doubling of distance to the nearest labelled building and declined for tall buildings whose rooflines often fell outside the frame. When images and records disagreed, predictions usually moved towards the supplied record. Released OpenFACADES floor annotations, generated with OpenStreetMap floor values as input, showed the same dependence: their agreement with the reference increased with building height, whereas that of image-only reruns decreased. The value of street-view imagery therefore depends on whether an attribute is visible and how well the place is already covered by existing data. Because machine-derived labels are often reused as references, these comparisons also bear on how urban datasets are documented and evaluated.
\end{abstract}

\vspace{8pt}
\noindent\textbf{Keywords:} street-view imagery; vision-language models; urban sensing; existing urban data; building attributes; data provenance

%% file: sec/1_intro.tex
\section{Introduction}

Street-view imagery is widely used to inventory buildings and infrastructure, monitor urban change, and estimate neighbourhood conditions \citep{biljecki2021svi,fan2025coverage,dai2025busstops,liu2025urbanchange}. Large language models are also being adopted for information extraction in urban analytics \citep{crooks2024information}. Their multimodal versions, vision-language models (VLMs), can be prompted to extract built-environment attributes from an image \citep{liang2025openfacades,jang2025auditing} or to rate how a place is perceived \citep{malekzadeh2025attractiveness}. The outputs of these models no longer stay within research benchmarks. OpenFACADES, for example, releases VLM-generated building attributes as an open dataset \citep{liang2025openfacades}, and such attributes can then serve as inputs or reference labels in later studies.

Cities that plan new image collection, however, rarely start without data. They already hold maps, administrative records, and observations from nearby locations, and they need to know where photographs add information beyond these sources. High predictive accuracy does not answer this question. A model may predict high house prices for a Manhattan street scene because it recognises the place, while neighbourhood statistics already give the same information. A municipal register may already locate many curb ramps, and visible road damage may be predictable from adjacent survey frames. In each case, an image-based model can perform well even though another data source already captures much of the relevant information.

Whether an image adds information depends on two properties. Visual legibility is the extent to which an attribute can be inferred from the photograph. Non-image predictability is the extent to which existing urban data can predict the same attribute without the photograph. The second property follows from the spatial structure of urban data: nearby places tend to be similar \citep{tobler1970growth}, and many attributes are already recorded in registers or open maps. Imagery should be most informative when the attribute is visible and existing urban data are sparse or weak.

Current evaluations say little about non-image predictability. Most street-view benchmarks evaluate images alone or combine them with text and location \citep{liang2025openfacades,yang2024commitments,liu2026citylens}, and location cues are known to change what multimodal models report \citep{kim2026geographic}. A coordinate-only baseline can show how much a model already knows about a location, but an urban analyst usually has more than coordinates, such as observations from nearby locations and administrative records. Our review of recent open-access studies that predict building attributes from street-level imagery found none that reported a predictor based only on location or neighbouring labels (Supplementary Section A).

We therefore compare image-based predictions with information already available for the same urban units, that is, the road frames, street corners, buildings, or statistical areas to which each label refers. The study covers infrastructure, building, and socioeconomic attributes from five public resources. For each task, we evaluate the image of each unit against a simple non-image alternative based on class prevalence, labels of nearby units, or a public record. We also replace images and add conflicting records to see which source the model follows. This last question also concerns public datasets, because some machine annotations are produced by a VLM that receives a map record together with the image.

The analysis asks three questions. First, which urban attributes gain useful predictive information from street-view imagery? Second, how does this contribution change with the density of existing labels around a place? Third, what happens when the image and an accompanying record disagree? Section 5 addresses the first two questions, and Section 6 the third.

The study contributes a paired design that compares street-view imagery with the non-image data available for the same urban units and that can be applied to other attributes and cities. With this design, we show that the added value of imagery varies with the local density of existing labels, so that the same attribute can call for new images in one district and not in another. We also show that supplied records shape both model predictions and released machine annotations, which affects how multimodal accuracy and derived datasets should be interpreted.

%% file: sec/2_related.tex
\section{Related Work}

Street-view imagery has become a general source for measuring buildings, roads, accessibility, environmental quality, and socioeconomic conditions \citep{biljecki2021svi}. Deep learning models trained on street-level images have estimated the demographic makeup of neighbourhoods \citep{gebru2017demographics}, house prices \citep{law2019houseprices}, and road damage \citep{maeda2018roaddamage}. Recent work has examined where imagery is available and how completely it captures building facades \citep{fan2025coverage}, and has applied it to transit amenities and urban change \citep{dai2025busstops,liu2025urbanchange}. Work on representation learning further shows that the spatial and temporal structure useful for one urban task may not suit another \citep{li2026stcl}.

A related line of work examines how acquisition and processing choices affect measures derived from street-view imagery. Image quality varies between platforms and captures \citep{hou2022sviquality}. Derived values also depend on road sampling intervals \citep{kim2021measurement}, viewing perspective \citep{biljecki2023sensitivity}, capture season \citep{zhao2025seasonal}, and weather \citep{ki2026weather}. \citet{ki2026weather} showed that restricting a sample to clear-weather images reduces weather-related measurement bias at the cost of spatial coverage. These studies ask how reliably an image measures an attribute. Our study asks whether, at a given place, the image is needed once existing data are taken into account.

Public datasets have widened the range of attributes that can be studied. OpenFACADES provides building imagery with machine-generated annotations \citep{liang2025openfacades}. Project Sidewalk records accessibility features through crowdsourcing \citep{saha2019projectsidewalk}, SVRDD covers road damage \citep{ren2024svrdd}, BuildingSense pairs building images with text descriptions for function classification \citep{su2026buildingsense}, and CityLens links street scenes with socioeconomic indicators \citep{liu2026citylens}. Multimodal foundation models can analyse the photograph together with coordinates or text \citep{yang2024commitments}. VLMs have been applied to built-environment auditing \citep{jang2025auditing}, walkability assessment with additional spatial data \citep{ki2025walkability}, and socioeconomic inference from street images \citep{li2024streetviewllm}. These resources and applications make urban prediction easier to benchmark, but most evaluations still compare models rather than the information sources available for the same place.

This distinction matters because a model can be accurate without using the intended signal. \citet{chen2024mmstar} found that many questions in VLM benchmarks can be answered without the image, and work on shortcut learning shows more generally that models often rely on cues other than the intended ones \citep{geirhos2020shortcut}. In urban sensing, an important alternative cue is location. Language models hold uneven geographic knowledge \citep{manvi2024geographic,moayeri2024worldbench}. \citet{kim2026geographic} further showed that adding coordinates or city and country names to a prompt shifts the ratings that multimodal models give to street scenes, even when the image is unchanged. Location information can therefore enter a prediction through the prompt as well as through the scene.

Spatial structure provides a second route. Urban observations are spatially correlated, so nearby labels can predict road or building conditions, and public registers may already contain the target attribute. The same structure affects validation. When nearby units share labels, random train-test splits overstate performance, and spatially blocked validation gives lower estimates that better reflect performance in new areas \citep{roberts2017crossvalidation,ploton2020spatial}. For buildings, height can be predicted from footprint geometry and the form of surrounding buildings \citep{milojevic2020heights}, but the building attributes recorded in OpenStreetMap are incomplete and their coverage varies between places \citep{biljecki2023buildingquality}. The relevant comparison for imagery is therefore local: how much does the target photograph add beyond the data available for the same place?

We reviewed open-access studies published since 2020 that predict objective building attributes from street-view imagery (Supplementary Section A). About half of the 33 retained studies reported a baseline, usually another image model or a method that uses an object in the scene for scale, and only four used spatial cross-validation. None reported a predictor based only on location or neighbouring labels. We therefore evaluate imagery against existing urban data on the same urban units.

%% file: sec/3_taxonomy.tex
\section{How the Urban Information Environment Shapes Image Value}
\label{sec:taxonomy}

We organise the study around visual legibility and non-image predictability. Neither is fixed for an attribute. Visual legibility changes with viewpoint and image quality, and non-image predictability changes with the age and coverage of the available records. Table~\ref{tab:taxonomy} lists, for each attribute, the image cue that might reveal it and the existing data that might predict it.

Visual legibility describes how clearly the target attribute can be inferred from a single street-level photograph at the resolution given to the model. Surface damage and curb ramps can appear directly in the frame. Floor count can be read only when enough of the elevation and roofline are visible. Architectural form and signage can indicate building type or function. Population and house price have no direct visual referent in a street scene, although the scene may still reveal the neighbourhood or city, and earlier image models have exploited such indirect cues \citep{gebru2017demographics,law2019houseprices}.

Non-image predictability describes how well existing urban data predict the value without the target photograph. Adjacent road frames may share surface condition, municipal programmes create spatial patterns in curb ramps, and neighbouring buildings often have similar form and height \citep{milojevic2020heights}. Area-level socioeconomic values also tend to vary smoothly across space.

\begin{table}[t]
  \centering
  \begin{threeparttable}
  \caption{Urban attributes and the image and non-image sources considered in
  the study.}
  \label{tab:taxonomy}
  \small
  \begin{tabularx}{\linewidth}{L{0.17\linewidth}L{0.24\linewidth}>{\raggedright\arraybackslash}X>{\raggedright\arraybackslash}X}
    \toprule
    {\tableheader Attribute} & {\tableheader Resource} &
    {\tableheader Potential image cue} & {\tableheader Existing urban data} \\
    \midrule
    Road damage & SVRDD\newline\citep{ren2024svrdd} & surface condition & nearby road observations \\
    \rowgray Curb ramp & Project Sidewalk\newline\citep{saha2019projectsidewalk} & corner geometry & municipal asset register \\
    Floor count & OpenFACADES\newline\citep{liang2025openfacades} & visible elevation and roofline & nearby OSM floor tags \\
    \rowgray Building type & OpenFACADES\newline\citep{liang2025openfacades} & facade form and use cues & local class prevalence \\
    Building function & BuildingSense\newline\citep{su2026buildingsense} & signage and entrances & nearby labels and place text \\
    \rowgray Population & CityLens\newline\citep{liu2026citylens} & indirect neighbourhood cues & neighbouring statistics \\
    House price & CityLens\newline\citep{liu2026citylens} & indirect neighbourhood cues & neighbouring statistics \\
    \bottomrule
  \end{tabularx}
  \begin{tablenotes}[flushleft]
    \footnotesize
    \item[] \textit{Note:} OSM = OpenStreetMap.
  \end{tablenotes}
  \end{threeparttable}
\end{table}

Both properties can differ between and within cities. A clear photograph may reveal floor count where mapped neighbours are sparse, while the same photograph adds little where a building register is well maintained. We therefore compare sources on the same urban units and examine how the result changes with the local density of existing labels.

Two comparisons are used throughout the paper. We call the city, coordinates, or released text that a resource gives to the model its task context. Within-model image gain is the change in a VLM's performance when the target image is added to this task context. Standalone image advantage is the difference between an image-only prediction and the strongest non-image alternative tested for that task. The first comparison shows whether a VLM uses the image. The second shows whether the image is more informative than a source that an urban analyst already has.

%% file: sec/4_protocol.tex
\Needspace{7\baselineskip}
\section{Study Design and Data}
\label{sec:protocol}

\subsection{Comparing sources on the same urban units}

For every urban unit, we compare predictions made from the target photograph with predictions that are available without it. The non-image alternative depends on what each resource provides: class prevalence, labels from nearby units, a municipal asset register, or neighbouring official statistics. We call predictors built from the labels of nearby units spatial predictors. The road- and building-label predictors ignore labels within 50 m of the target to reduce dependence on near-duplicate building views or adjacent road observations. Wider exclusion radii are examined in the supplement. Image and non-image predictions are always compared on the same set of units.

We separate task context from existing urban data (Table~\ref{tab:context-levels}). Task context is given directly to the VLM as part of the prompt. Existing urban data are labels, registers, and statistics that an analyst can use without any VLM. The standalone comparison uses the strongest tested non-image source for each task, so an image advantage cannot arise from selecting a weaker tested baseline. Supplementary Table S1 lists the alternatives considered for each resource, and the companion archive reports results for all of them and for the spatial settings of the nearby-label predictors.

\begin{table}[t]
  \caption{Non-image sources used in the comparisons.}
  \label{tab:context-levels}
  \centering
  \small
  \begin{tabular}{lll}
    \toprule
    {\tableheader Source} & {\tableheader Information supplied} &
    {\tableheader Role in the study} \\
    \midrule
    Task context & city, coordinates, or released text & VLM without the image \\
    \rowgray Nearby observations & labels from other urban units & spatial alternative \\
    Public record & register or official statistics & administrative alternative \\
    \bottomrule
  \end{tabular}
\end{table}

\subsection{Paired source comparisons}

Where the resource permits, each unit is evaluated under three conditions: image only, task context only, and image plus task context. Within-model image gain compares the last two conditions, and standalone image advantage compares the image-only condition with the non-image alternative. The CityLens tasks depart from this design in several respects, which are described in Section~\ref{sec:datasets}.

\begin{figure}[t]
  \centering
  \includegraphics[width=\linewidth]{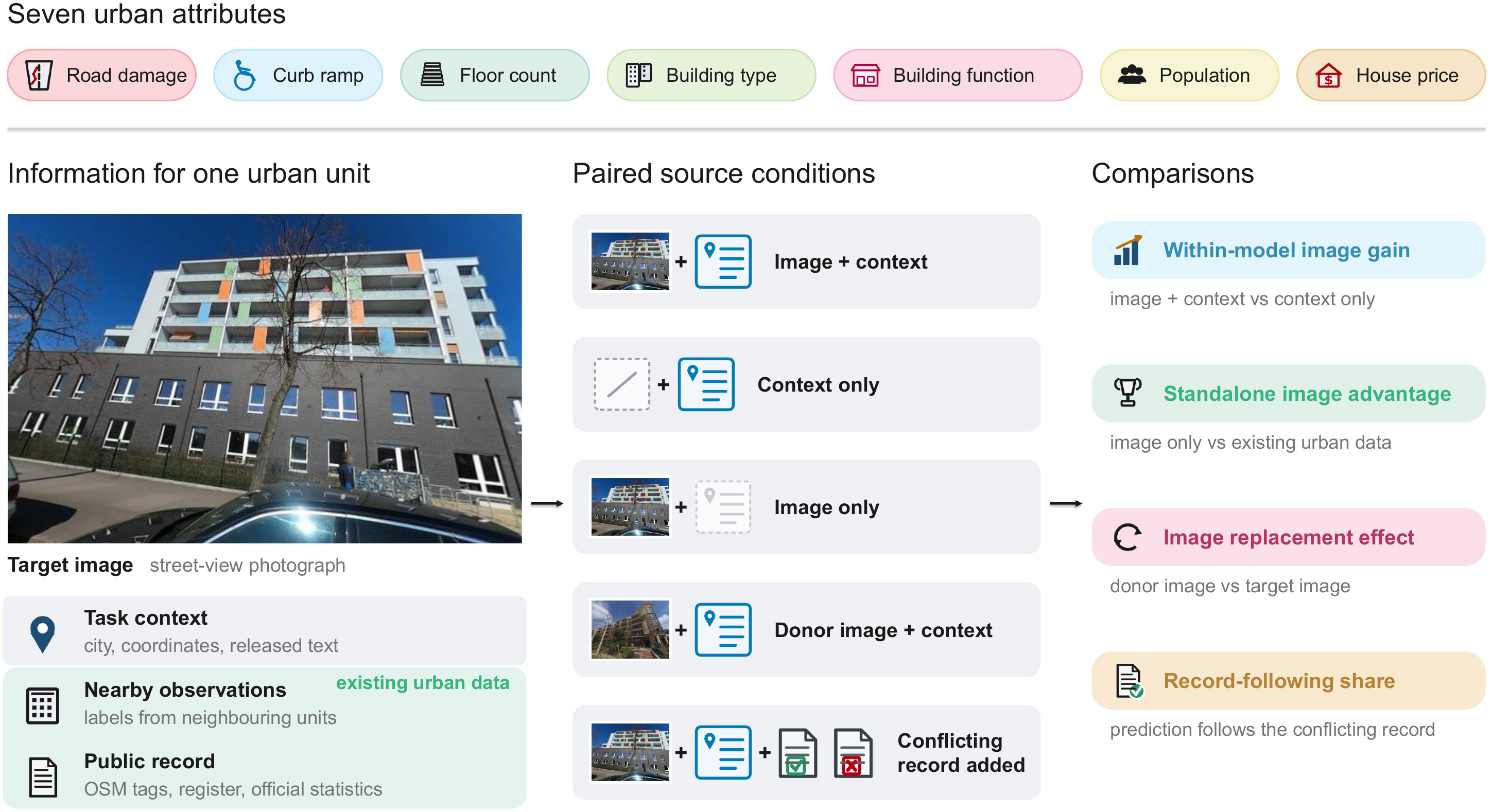}
  \caption{Study design for comparing street-view imagery with existing urban
  data. The figure links the information available for an urban unit to paired
  source conditions, non-image alternatives, and the comparisons used to
  identify image and record influence.}
  \label{fig:framework}
\end{figure}

We then change one information source at a time while keeping the evaluated unit fixed (Figure~\ref{fig:framework}). In the image-replacement condition, the target photograph is replaced by a donor image from another unit, and the task context is kept. The image replacement effect is the resulting loss of accuracy. In the conflicting-record condition, a register-style statement is added that disagrees with the reference value, that is, the label used for scoring. The record-following share is the proportion of predictions that follow the supplied value. For BuildingSense, the text descriptions are also rerun after removing nearby points of interest or replacing the address. Prompt specifications are given in Supplementary Section I.

\subsection{Public resources and urban settings}
\label{sec:datasets}

The infrastructure cases are road-damage images from SVRDD in Beijing \citep{ren2024svrdd} and curb-ramp observations from Project Sidewalk in Seattle \citep{saha2019projectsidewalk}. We sampled 300 road frames without replacement, taking 60 from each of five districts. The curb-ramp sample contains 300 locations, balanced between ramp and no-ramp labels, drawn in random order from clusters with at least two observations and more agreements than disagreements. Road images are compared with labelled frames nearby, and curb-ramp images with the asset register of the Seattle Department of Transportation (SDOT).

The building cases use OpenFACADES for floor count and building type \citep{liang2025openfacades} and BuildingSense for building function \citep{su2026buildingsense}. The spatial floor-count analysis uses 448 buildings in seven metropolitan areas, sampled by height with one image per building: 150 with 1--4 floors, 149 with 5--14, and 149 with 15 or more. Paired context and source-conflict experiments use a core sample of 150 buildings, 50 per height group; 97 have mapped building-type references. Floor-count references are the OpenStreetMap (OSM)-derived single-attribute values retained from the OpenFACADES release. We retrieved additional OSM attributes for the released building identifiers, including the building tags used as building-type references, and use the tags of other nearby buildings as the non-image alternative. The VLM-generated multi-attribute annotations released with OpenFACADES are not used as references; Section 6.3 examines how they were produced. BuildingSense contributes 200 New York buildings randomly sampled from 979 eligible image--description pairs, without class balancing. Each description contains the building's address, dates, height, and nearby points of interest.

The socioeconomic cases use the CityLens population and house-price tasks \citep{liu2026citylens}. We sampled areas in random order from those with available image bundles. The population sample contains 150 areas in eleven cities, each with ten street views and one satellite image, and the house-price sample contains 120 tracts, 60 each in New York and Chicago, with ten street views and no satellite image per tract. CityLens reports each indicator as a score from 0.0 to 9.9. In the released task files, this score equals the percentile rank of an area's value among all areas of the task, pooled across cities, to within 0.1. For the US house-price tracts, CityLens takes values from the ZIP-code-level Zillow Home Value Index and maps them onto census tracts \citep{liu2026citylens}, so tracts that fall in the same ZIP code receive the same value.

\begin{figure}[t]
  \centering
  \includegraphics[width=\linewidth]{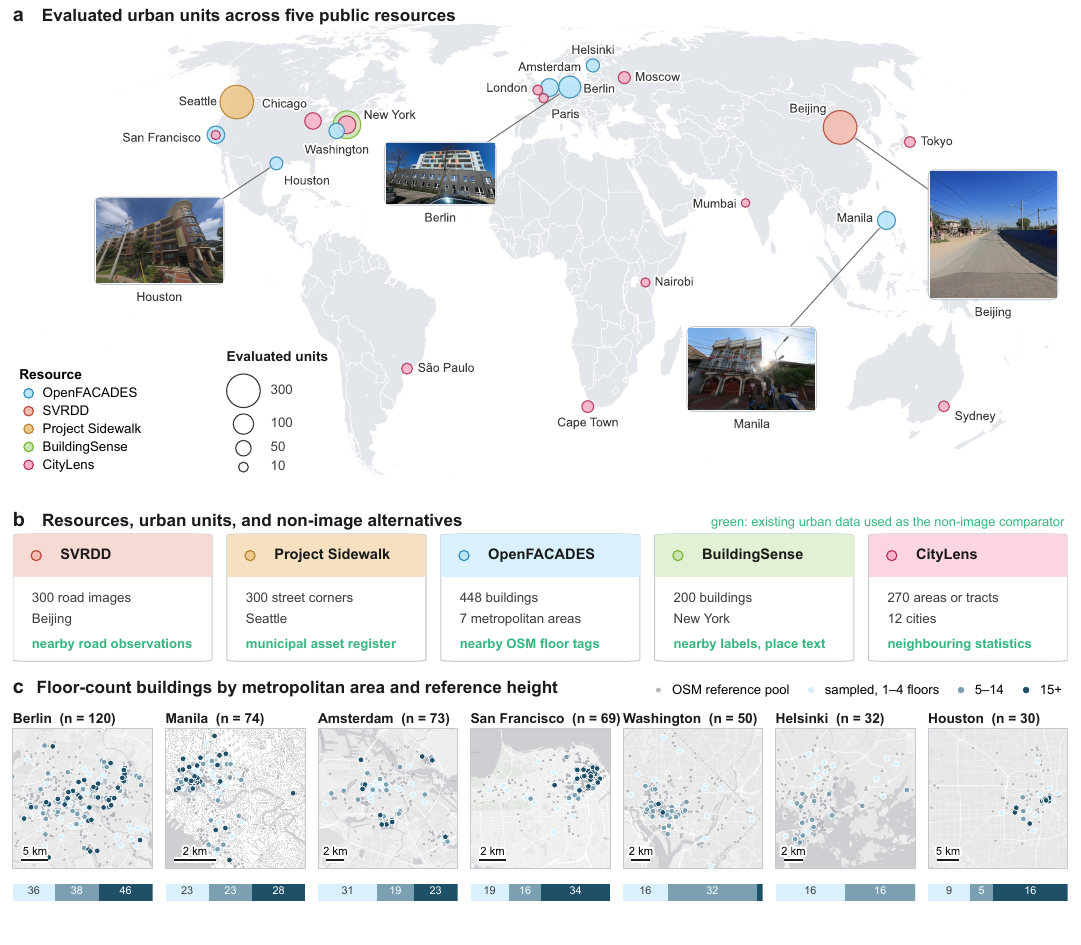}
  \caption{Study resources, urban units, and geographic coverage.
  \textbf{(a)}~Evaluated urban units in the five public resources. Symbol
  area is proportional to the number of units in each city, and the
  photographs show example units from OpenFACADES (Mapillary, CC BY-SA) and
  SVRDD. \textbf{(b)}~Unit of analysis, evaluated units, coverage, and, in
  green, the existing urban source used as the non-image comparator for each
  resource.
  \textbf{(c)}~Floor-count buildings in the seven metropolitan areas, shaded
  by reference height, over the OpenStreetMap reference pool in grey; the
  strip under each map gives the count per height stratum. Basemap: Esri
  World Light Gray Canvas.}
  \label{fig:study-area}
\end{figure}

CityLens requires several departures from the general design. Its released prompt names the city, so the image-only condition removes the city name. Its location-only prompt replaces the images with centroid coordinates, so the within-model contrast removes coordinates while adding imagery. The non-image alternative is the median score of the five nearest other areas, without an exclusion radius; Supplementary Table S12 reports a VLM that receives these statistics without an image. Image replacements exchange the full bundle with one from another city while keeping the target prompt, and their effect is measured as the share of predictions within one point of the reference. Figure~\ref{fig:study-area} shows the geographic coverage, urban units, and non-image source of each resource. Sample construction, spatial matching, and temporal alignment are described in the supplement.

\paragraph{Models and evaluation.}
We evaluate GPT-4o-mini, Gemini 2.5 Flash Lite, and Qwen3.5 Flash under every source condition that a resource allows, using fixed prompts at temperature zero. These lightweight models span three providers and make repeated source-condition comparisons practical. Classification tasks are scored with accuracy and class-balanced measures. Road damage has six binary damage classes and is scored with balanced accuracy, the mean of sensitivity and specificity averaged over the classes, because always predicting no damage already reaches a mean per-class accuracy of 75.3\%. CityLens source comparisons use mean absolute error (MAE) on the released score scale. Uncertainty intervals are computed over paired urban units. Supplementary Sections I and J give prompt details, model settings, response validity, and full estimates.

%% file: sec/5_study1.tex
\section{Where Street-View Imagery Adds Information}
\label{sec:study1}

\begin{figure}[t]
  \centering
  \includegraphics[width=\linewidth]{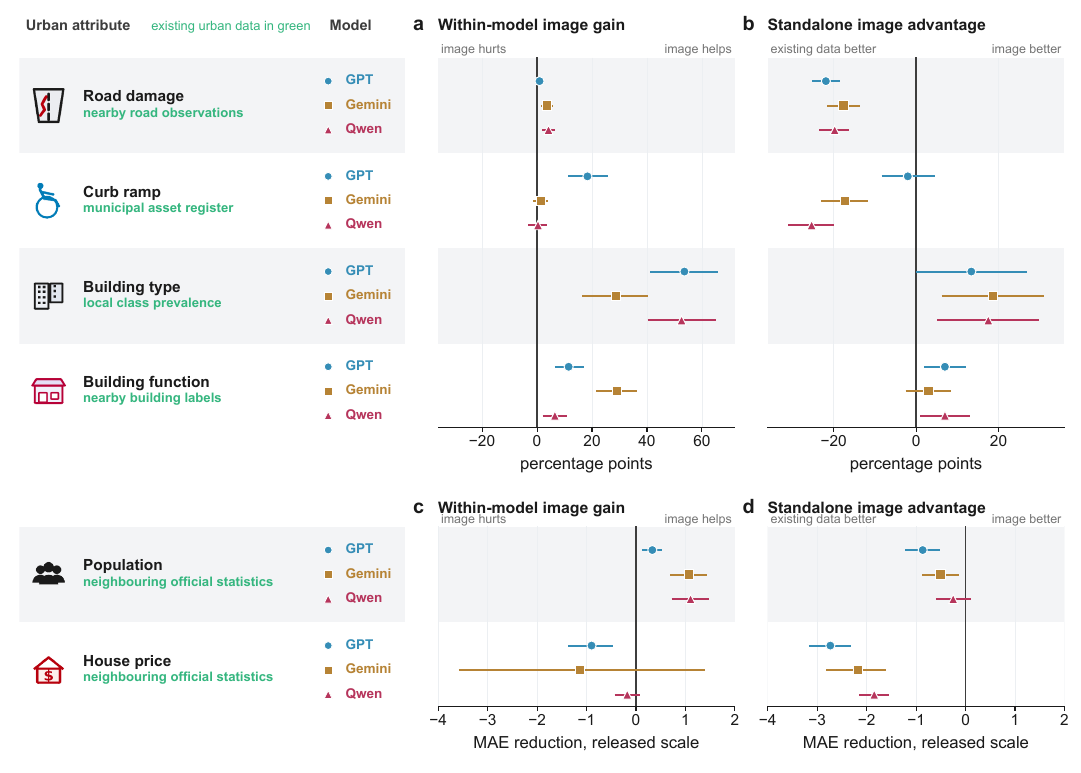}
  \caption{Image contribution across the evaluated urban attributes.
  \textbf{(a)}~Change when the target image is added to the same VLM's task
  context. \textbf{(b)}~Difference between the image-only prediction and the strongest tested non-image source. \textbf{(c)}~CityLens MAE reduction from the location-only prompt, which includes centroid coordinates, to the released image prompt. \textbf{(d)}~CityLens MAE reduction from the median of neighbouring official statistics to image-only predictions without the city name. Road damage is scored with balanced accuracy. Positive values favour the image condition. Points
  identify models and horizontal lines show paired uncertainty intervals.
  Green text names the existing urban source used in (b) and (d).}
  \label{fig:cross-task}
\end{figure}

Figure~\ref{fig:cross-task} reports the source comparisons for six of the seven attributes, using the CityLens conditions specified in Section~\ref{sec:datasets}. Table~\ref{tab:main-results} gives the scores and within-model image gains. Floor count is examined separately in Section 5.2 because its image value depends on building height and data coverage. Across these attributes, image value varied more between attributes than between models, which points to properties of the attributes and their data rather than of a particular VLM.

\begin{table}[htbp]
  \centering
  \begin{threeparttable}
  \caption{Image-only scores, strongest tested non-image sources, and
  within-model image gains. Accuracy scores are percentages; MAE uses the
  released CityLens score scale. GPT, Gemini, and Qwen denote the three
  models specified in Section 4.3.}
  \label{tab:main-results}
  \small
  \setlength{\tabcolsep}{3pt}
  \begin{tabularx}{\linewidth}{@{}L{0.135\linewidth}L{0.125\linewidth}XrrrR{0.15\linewidth}@{}}
    \toprule
    {\tableheader Attribute} & {\tableheader Metric} &
    {\tableheader Non-image source: score} &
    \multicolumn{3}{c}{\tableheader Image only} &
    {\tableheader Image gain range} \\
    \cmidrule(lr){4-6}
    & & & {\tableheader GPT} & {\tableheader Gemini} & {\tableheader Qwen} & \\
    \midrule
    Road damage & Balanced accuracy & Nearby frames: 72.3 & 50.4 & 54.7 & 52.6 & $+0.9$ to $+4.2$ \\
    \rowgray Curb ramp & Accuracy & SDOT register: 78.0 & 76.0 & 60.7 & 52.7 & $+0.3$ to $+18.3$ \\
    Floors, 1--4 & Within 1 floor & Nearby tags: 59.3 & 92.0 & 88.7 & 86.0 & $+56.0$ to $+90.0$ \\
    \rowgray Floors, 5--14 & Within 1 floor & Nearby tags: 43.2--43.9 & 40.3 & 64.2 & 54.7 & $+10.0$ to $+34.0$ \\
    Floors, 15+ & Within 1 floor & Nearby tags: 17.4--17.7 & 3.4 & 21.5 & 18.4 & $+2.0$ to $+16.0$ \\
    \rowgray Building type & Accuracy & Modal class: 58.8 & 72.2 & 77.3 & 76.3 & $+28.9$ to $+53.6$ \\
    Building function & Accuracy & Nearby labels: 56.0 & 63.0 & 59.0 & 63.0 & $+6.5$ to $+29.0$ \\
    \rowgray Population & MAE & Nearby statistics: 1.78 & 2.65 & 2.28 & 2.03 & $+0.33$ to $+1.10^{\dagger}$ \\
    House price & MAE & Nearby statistics: 0.17--0.24 & 2.97 & 2.35 & 2.08 & $-1.13$ to $-0.18^{\dagger}$ \\
    \bottomrule
  \end{tabularx}
  \begin{tablenotes}[flushleft]
    \footnotesize
    \item[] Non-image scores use each model's valid image-only sample;
    ranges reflect differences in these samples. Image gain compares image
    plus task context with context alone on a separate paired sample. It is
    measured in percentage points for accuracy and in MAE reduction for
    CityLens; positive values favour the image condition. For road damage,
    balanced accuracy is the mean of sensitivity and specificity, averaged
    over six binary damage classes; always predicting no damage scores 50.0.
    \item[] Floor scores use the 448-building spatial sample; floor gains
    use the 150-building core sample. CityLens image-only runs omit city
    names, and the non-image score is the median score of the five nearest
    areas. Gemini house-price image-only scores use 43 valid tracts (matched
    non-image MAE 0.17); its within-model gain uses 10 valid pairs.
    \item[] $\dagger$ CityLens gains compare the released image prompt with
    the location-only prompt, which also supplies centroid coordinates.
    They do not isolate the effect of adding an image to identical context.
  \end{tablenotes}
  \end{threeparttable}
\end{table}

\subsection{Image value differs by attribute}

For road damage, curb ramps, population, and house price, the image was not the most informative source. Nearby labelled frames reached a balanced accuracy of 72.3\% for road damage, whereas the image-only models reached 50.4\% to 54.7\%, close to the 50\% obtained by always predicting no damage (Table~\ref{tab:main-results}). The spatial predictor lost accuracy as the exclusion radius widened (Supplementary Table S2), which indicates that damage labels are clustered along the surveyed roads. The images performed poorly for a different reason: the models rarely reported damage, and their positive recall ranged from 4.8\% to 23.7\% (Supplementary Table S3). Road damage is visible in principle, so this result reflects the detection ability of the tested VLMs as much as the information contained in the photographs.

Curb ramps and population followed the same pattern for other reasons. Seattle maintains a municipal register of curb ramps, and this register matched or exceeded the image-only predictions. Population, like other area-level indicators, tends to vary smoothly across neighbouring areas, and all image-only runs without the city name had higher MAE than the median of neighbouring statistics (2.03 to 2.65 against 1.78; Table~\ref{tab:main-results}), although the paired interval for Qwen3.5 Flash included zero (Figure~\ref{fig:cross-task}d). Earlier studies have predicted such socioeconomic indicators from street-level images \citep{gebru2017demographics,law2019houseprices,li2024streetviewllm}. Our comparison adds a different reference point: for a given area, the statistics of its neighbours already carry much of the same information.

House price showed a much larger margin. The neighbour median reached an MAE of 0.17 to 0.24, against 2.08 to 2.97 for the images. Much of this margin comes from how the reference was constructed. Because CityLens assigns ZIP-code values to US census tracts (Section~\ref{sec:datasets}), 68\% of the sampled tracts shared their exact value with at least one of their five nearest CityLens tracts, and the neighbour median equalled the target score for 46\% of tracts. The house-price comparison is therefore dominated by the spatial resolution of the reference, and it cannot show whether images capture price variation within a ZIP code.

The pattern reversed for building attributes that are visible in the scene. All three image-only models exceeded the non-image alternative for building type, and the difference remained large under class-balanced measures. Their balanced accuracy ranged from 63.1\% to 76.1\%, against 16.7\% for the modal class and 25.2\% for the spatial predictor (Supplementary Table S6). Neighbouring labels were a poor guide to building type. The spatial predictor was less accurate than always predicting the most common class (27.8\% against 58.8\%), whereas the facade shows the type directly. Building function also benefited from imagery. In overall accuracy the difference was smaller and depended on the model, most likely because most buildings in the sample are residential and nearby labels already predict this majority class. Under class-balanced measures, all three models exceeded the nearby-label predictor (Supplementary Table S9).

The two comparisons did not always lead to the same conclusion. Adding a curb-ramp image to the location context improved some VLM answers, yet the image-only predictions did not improve on the municipal register. Road damage showed the same distinction: adding the image to the district name and coordinates raised balanced accuracy by 0.9 to 4.2 points, but image-only predictions remained 17.6 to 21.9 points below nearby frames. When task context is held fixed, a positive within-model gain shows that the model uses the image. It does not show that the image is the most informative source available.

\subsection{Floor-count value changes with building height and data coverage}
\label{sec:floorcase}

\begin{figure}[t]
  \centering
  \includegraphics[width=\linewidth]{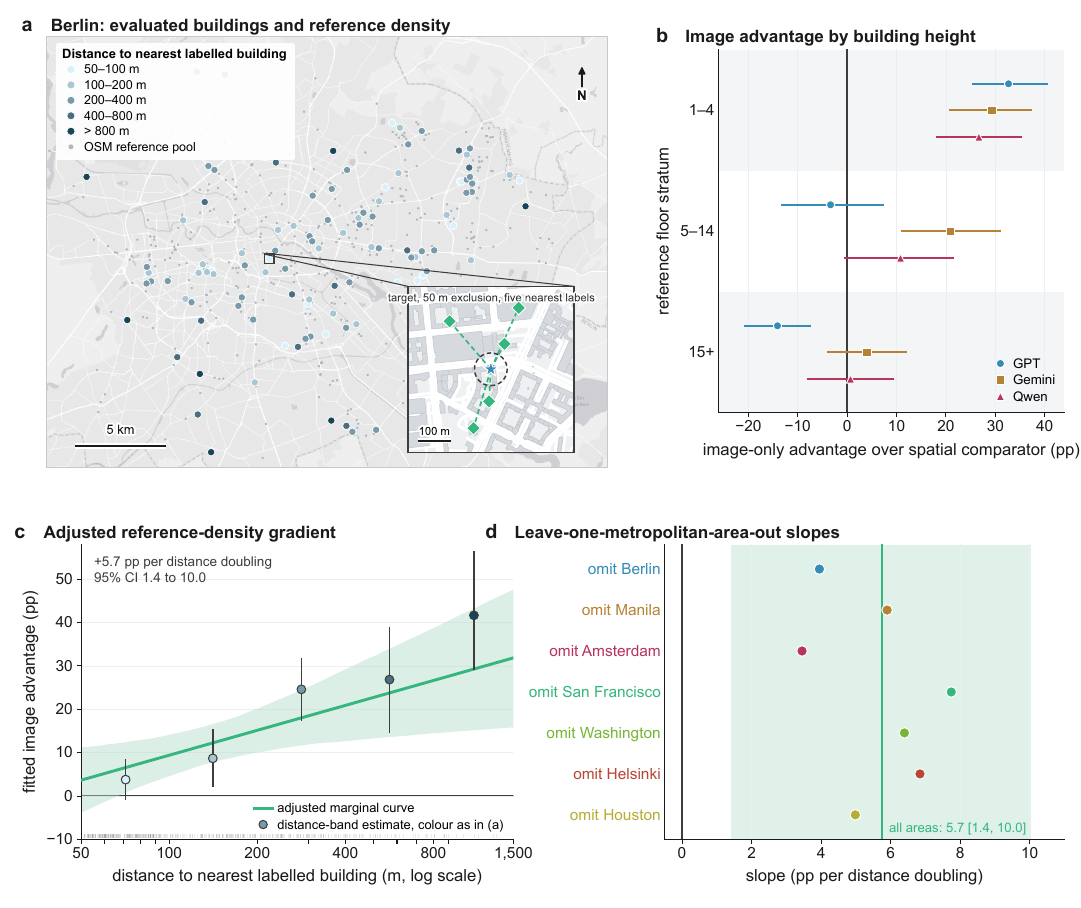}
  \caption{Spatial variation in the image contribution to floor-count
  prediction. \textbf{(a)}~Evaluated buildings in Berlin, shaded by the
  distance to the nearest labelled building in the OpenStreetMap reference
  pool (grey). The inset shows one target with its 50\,m exclusion ring and
  its five nearest labelled neighbours over OpenStreetMap building
  footprints. \textbf{(b)}~Image advantage by building-height group and model.
  \textbf{(c)}~Adjusted image advantage as the nearest labelled building
  becomes more distant; the band estimates use the colours of (a).
  \textbf{(d)}~The distance relationship after omitting each metropolitan
  area in turn. Basemap: Esri World Light Gray Canvas; footprints and streets
  from OpenStreetMap contributors.}
  \label{fig:floor-decomposition}
\end{figure}

Floor count allows us to examine, within one task, how image value changes with visual legibility and with local data coverage. The target buildings span seven metropolitan areas where OpenStreetMap floor tags are unevenly available (Figure~\ref{fig:study-area}c), a pattern also reported in global assessments of OpenStreetMap building attributes \citep{biljecki2023buildingquality}. For each target, the spatial predictor takes the median floor count of its five nearest labelled buildings outside the 50 m exclusion radius (Figure~\ref{fig:floor-decomposition}a). Using five neighbours reduces dependence on any single neighbouring label; sensitivity results for the radius and neighbour count appear in Supplementary Section C. A prediction counts as correct when it is within one floor of the reference, because OpenStreetMap level counts exclude roof storeys, which can appear as an extra floor in a photograph; exact-match results are reported in the supplement.

Building height determined how much of the attribute the photograph showed. Images were most helpful for low-rise buildings, whose elevation and roofline were usually visible, and for buildings of one to four floors all three models exceeded the spatial predictor by roughly 27 to 33 percentage points (Figure~\ref{fig:floor-decomposition}b). The advantage became small or negative for tall buildings, whose upper floors often extended beyond the frame. Our inspection of sampled high-rise images found that most omitted the roofline or were taken at a steep angle (Supplementary Section C).

Local data coverage had a separate effect. Figure~\ref{fig:floor-decomposition}c plots adjusted image advantage against nearest-label distance on a logarithmic horizontal axis. The image advantage increased as mapped neighbours became more distant. It was about five percentage points when the nearest labelled building was within 100 m and more than thirty points beyond 400 m. After adjusting for metropolitan area, building height, camera distance, and image year, the advantage increased by 5.7 percentage points per doubling of distance (95\% CI 1.4 to 10.0). Over the 50 m to 1.5 km range shown in Figure~\ref{fig:floor-decomposition}c, this slope corresponds to a fitted increase of about 28 percentage points. The slope remained positive when each metropolitan area was omitted in turn (Figure~\ref{fig:floor-decomposition}d; Supplementary Table S7). A replication with a supervised image classifier showed the same direction. Under geographically blocked splits, which leave fewer labelled buildings near each test building, the image advantage widened from about 12 to 21 percentage points (Supplementary Section D).

The explanation lies in how the two sources work. The spatial predictor depends on the similarity between a building and its mapped neighbours, and this similarity weakens with distance, as studies of spatial validation have shown for other mapping tasks \citep{ploton2020spatial}. The photograph does not depend on neighbours. Its relative value therefore rises where the map is sparse, even when the attribute is equally legible in both places. For image collection, this means that the same attribute can call for new imagery in one district and not in another.

\subsection{Place descriptions reduce the incremental value of imagery}
\label{sec:bsense-text}

BuildingSense descriptions contain the address, dates, height, and nearby points of interest of each building, so the text alone can reveal both the place and its likely use. Adding nearby points of interest raised context-only accuracy by 18.5 to 23.1 percentage points across the three models, whereas replacing the address changed little (Supplementary Table S10). The useful information in the text was therefore the description of the building's surroundings rather than its address. Images were most helpful for buildings whose description named no point of interest. This case resembles the problem identified by \citet{chen2024mmstar} in general VLM benchmarks, where many items can be answered from text alone. In BuildingSense, a high score in the image-plus-text condition may reflect the description as much as the building image. Class-balanced results appear in Supplementary Section F.

%% file: sec/6_study2.tex
\section{When Images and Records Disagree}
\label{sec:study2}

\begin{figure}[t]
  \centering
  \includegraphics[width=\linewidth]{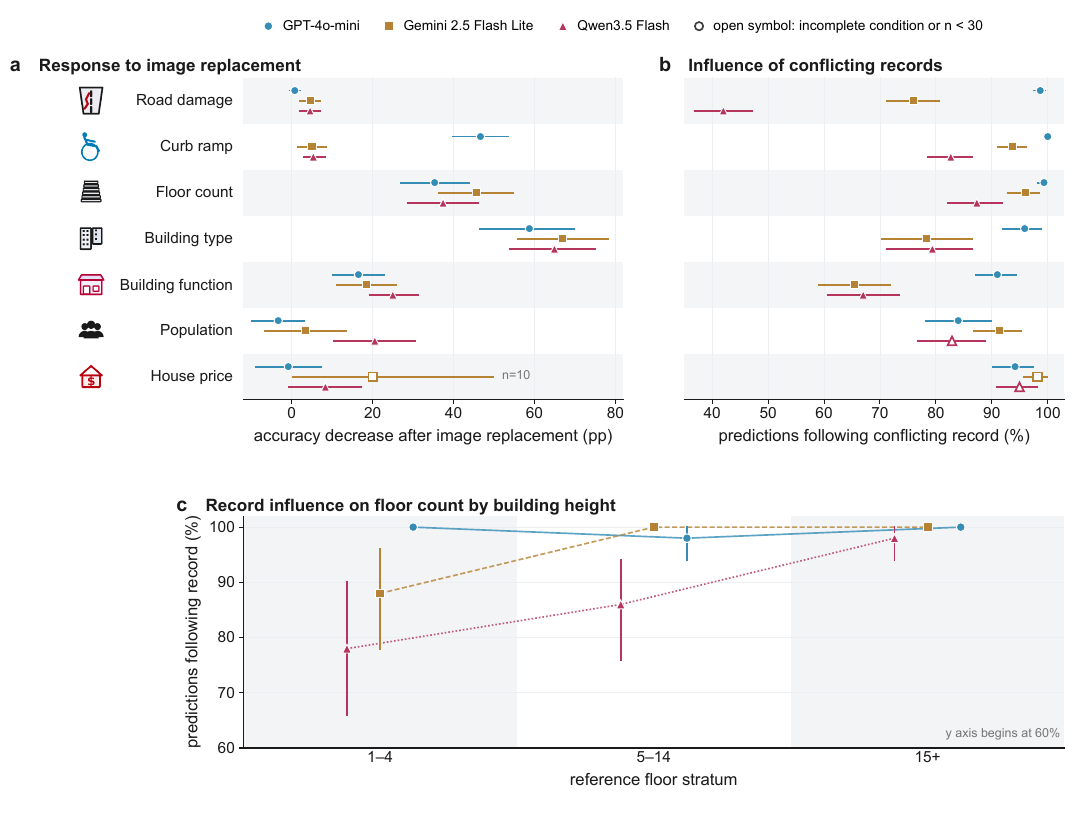}
  \caption{Predictions when image and record sources disagree.
  \textbf{(a)}~Change after replacing the target photograph while retaining
  its task context; road damage is scored with balanced accuracy.
  \textbf{(b)}~Share of predictions that follow a conflicting
  record. \textbf{(c)}~Record influence on floor-count predictions by building
  height. Points identify models and lines show paired uncertainty intervals.}
  \label{fig:conflict}
\end{figure}

The comparisons above do not show which source a model relies on when it receives several. Source conflicts address this question. A model that responds to the scene should change its answer when the image is replaced, and a model that follows the accompanying information should change its answer when a conflicting record is added. Figure~\ref{fig:conflict}a reports the accuracy loss after image replacement, and Figure~\ref{fig:conflict}b the share of predictions that follow a conflicting record.

\subsection{Image replacement effects differ across attributes}

Replacing the photograph had the largest effect on building type and floor count, where accuracy fell by 59 to 67 and 35 to 46 percentage points, respectively (Supplementary Table S11). The new prediction usually moved towards the donor building, as expected for attributes that are read from the visible form of a building. Road-damage predictions changed little (at most 4.6 points) even when the donor frame carried a different damage label. This is consistent with Section 5.1: models that rarely report damage give much the same answer whatever road they are shown. Curb ramps differed by model. Accuracy fell by 46.7 points for GPT-4o-mini but by about five points for the other two models. GPT-4o-mini was also the only model whose image-only curb-ramp predictions came close to the municipal register (Figure~\ref{fig:cross-task}b), so it appears to be the one model that relied on the corner itself.

For population and house price, donor image bundles came from another city while the prompt kept the target city name, and replacement had mixed effects. Accuracy fell clearly only for Qwen3.5 Flash on population, and some predictions moved towards the donor area, so the images carried some information about the place. The small population effects for GPT-4o-mini and Gemini, however, are consistent with these models relying on the city name more than on the images. In either case, the images did not make predictions more accurate than neighbouring official statistics.

\subsection{Supplied records often shape the prediction}

When a register-style statement contradicted the reference, predictions usually moved towards the supplied value (Figure~\ref{fig:conflict}b). This happened in infrastructure, building, and socioeconomic tasks, although the strength of the effect varied by model. For road damage, for example, GPT-4o-mini followed the record for almost every unit, whereas Qwen3.5 Flash followed it for 42\% of units. \citet{kim2026geographic} reported a related effect in perception tasks, where adding location information to the prompt lowered ratings of wealth and beauty for African cities and raised ratings for cities in Asia and North America. Our results show that supplied text has a similar influence on objective attributes, for which a reference value exists and the shift can be checked. A correct multimodal prediction may therefore reflect the accompanying record as much as the photograph.

Floor count shows how visual difficulty changes this relationship. Figure~\ref{fig:conflict}c plots the record-following share for each building-height group and model. Gemini and Qwen followed the supplied record most often for tall buildings, where the roofline was rarely visible, and less consistently for low-rise buildings. GPT-4o-mini followed it at every height. For the first two models, reliance on the record thus increased as the image offered less usable visual evidence. Supplementary Section H compares predictions under correct and incorrect records.

\subsection{Released OpenFACADES annotations combine image and record}

\begin{figure}[t]
  \centering
  \includegraphics[width=\linewidth]{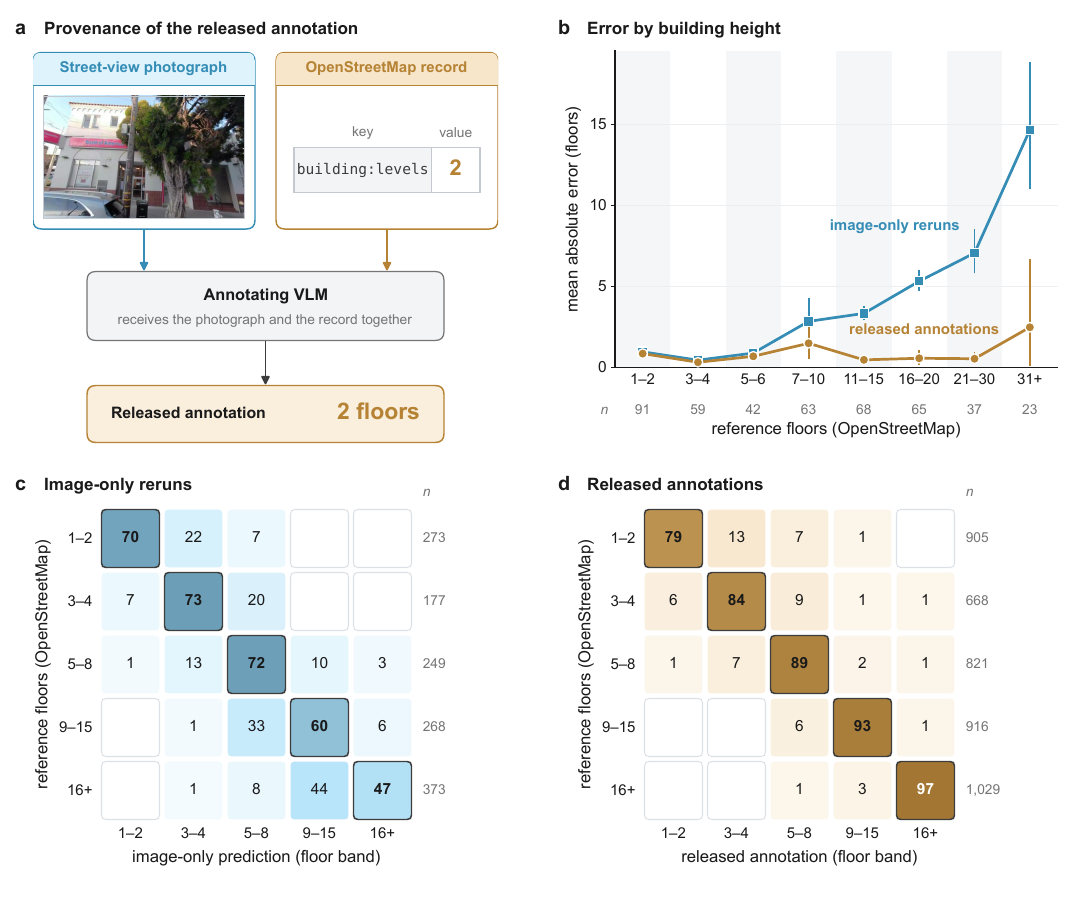}
  \caption{Influence of OpenStreetMap reference values on the released
  OpenFACADES annotations. \textbf{(a)}~The released workflow supplied the
  image and a floor-count record to the annotating model. \textbf{(b)}~Mean
  absolute error by building height for image-only reruns and released
  annotations on matched buildings; vertical lines show bootstrap 95\%
  intervals, and $n$ is the number of buildings in each height band.
  \textbf{(c,d)}~Floor-band confusion matrices using all available
  image-only predictions and released annotations, respectively; cells give
  the percentage of each reference band, and $n$ is the number of
  predictions or annotations in the band.}
  \label{fig:forensics}
\end{figure}

The same reliance on records can affect machine annotations when the annotating model receives a record. In the released OpenFACADES workflow, a commercial VLM received the street image together with the OpenStreetMap floor value before generating each annotation (\citealp{liang2025openfacades}; Figure~\ref{fig:forensics}a). The released floor labels may therefore draw on the record as well as on the image.

Their error pattern supports this reading. We scored the released annotations and our image-only reruns against the OSM-derived floor references. On matched buildings, image-only error rose steeply with building height, as expected when upper floors leave the frame, whereas released-annotation error stayed low up to 30 floors (Figure~\ref{fig:forensics}b). The floor-band matrices, which use all available predictions, show the same contrast (Figure~\ref{fig:forensics}c,d): only 47\% of image-only predictions for buildings of 16 or more floors fell in the correct band, whereas the share of released annotations in the correct band rose from 79\% for one- and two-floor buildings to 97\% for buildings of 16 floors or more. Because the annotating model had seen the OpenStreetMap value, this close agreement for tall buildings is consistent with the model copying the supplied value. The record-conflict experiment in Section 6.2 showed the same behaviour for tall buildings. Some released captions also named OpenStreetMap as the source of a value (Supplementary Section H).

The released labels are therefore annotations informed by reference data rather than independent image-only labels. When they are used to evaluate image-only models, agreement with them partly measures agreement with OpenStreetMap. Supplementary Section G gives the error-pattern analysis.

%% file: sec/7_study3.tex
\subsection{Patterns across models}
\label{sec:study3}
The main attribute-level pattern recurred across the three VLMs. Existing urban data were as strong as or stronger than the images for road damage, curb ramps, population, and house price, whereas images were stronger for building type and, under class-balanced measures, building function. For all three models, floor-count gains were concentrated in low-rise buildings.

Model behaviour still differed within tasks. On curb ramps, image replacement affected GPT-4o-mini much more than the other two models, and responses to the socioeconomic tasks were less consistent. The share of predictions following a conflicting record also varied by model, and behaviour on one attribute did not reliably predict behaviour on another. Detailed model results and incomplete responses appear in Supplementary Section J.

%% file: sec/8_discussion.tex
\section{Discussion}
\label{sec:discussion}

Our results answer the three questions posed in the Introduction as follows. Imagery added information for attributes that are read from building form, and it added little where nearby observations or public records already described the target. Within the floor-count task, its value depended on place as well as on the attribute, growing where mapped neighbours were distant. When image and record disagreed, the record usually prevailed, particularly when the photograph showed little of the target, and the same dependence appears in released machine annotations. Below we discuss why these patterns arise, and then what they mean for agencies that collect imagery (Section 7.2) and for those who evaluate models or release derived datasets (Section 7.3).

\subsection{Why image value varies}

Image value reflected both what the camera could see and what existing data already recorded. Visual legibility set an upper limit. Population and house price have no direct visual referent. The median of neighbouring statistics had lower MAE than every image-only run without the city name for both attributes (Table~\ref{tab:main-results}), consistent with the smooth spatial variation of such values \citep{tobler1970growth} and, for house price, with a reference assigned at ZIP-code level. Legibility alone was not sufficient, however. Road damage can appear in the frame, but the tested VLMs rarely reported it. Supervised detectors trained on labelled road images are the usual approach to this task \citep{maeda2018roaddamage,ren2024svrdd}, and a general-purpose VLM used without such training may miss damage that a specialised detector would find. Where the attribute was legible and the model could read it, as for building type and low-rise floor count, images added information that neighbouring labels did not provide.

The floor-count gradient adds a spatial dimension. The photographs came from the same resource and addressed the same attribute, yet their advantage increased where mapped neighbours were farther away. The coverage of OpenStreetMap building attributes differs between and within cities \citep{biljecki2023buildingquality}, so the benefit of imagery is also uneven across a city. A single benchmark score averages over these conditions and hides the variation. Reporting results by local label density, or validating with spatial blocks as recommended for other spatial prediction tasks \citep{roberts2017crossvalidation}, would make it visible.

Visual legibility also varied within the floor-count task. Low-rise facades often showed the full elevation, whereas tall buildings frequently extended beyond the frame. With building-height group and the other covariates held fixed, the image advantage rose by 9.1 percentage points per doubling of camera-to-building distance (95\% interval 1.2 to 16.9; Supplementary Table S7), consistent with more distant views keeping the roofline in the frame. Collecting imagery for tall buildings therefore requires viewpoints that capture the roofline, in addition to broad geographic coverage \citep{fan2025coverage}.

Two findings from the wider literature suggest a possible explanation for record-following. Visual question answering models can rely on language priors when questions suggest likely answers \citep{goyal2017vqa}. Large language models can also accept coherent and convincing external text that conflicts with their parametric knowledge \citep{xie2024knowledgeconflicts}. A register-style statement supplies a plausible value in the format of the requested answer. Visual evidence may compete with this value when the whole facade is visible, but may be less informative when upper floors lie outside the frame. The rise in record-following with building height for Gemini and Qwen (Figure~\ref{fig:conflict}c) is consistent with this account.

The question in the title, whether a model sees the city or recognises the place, has a task-dependent answer. For building form, image replacement moved predictions towards the donor building, so the models responded to what they saw. For population, two of the three models appeared to recognise the place rather than read the scene: when the images were swapped for another city's while the city name stayed in the prompt, their estimates kept similar accuracy. The median of neighbouring statistics also predicted both socioeconomic attributes with lower error than the image-only runs. For building function, the informative part of the text was the description of the surroundings, while the address mattered little. This differs from the perception tasks studied by \citet{kim2026geographic}, where adding location information to the same image shifted ratings. For objective attributes, the influence of the accompanying text appears to depend on how much task-relevant information it carries.

\subsection{Implications for image collection and mapping}

For agencies that plan image collection, the results suggest comparing sources before large-scale acquisition. In Seattle, a current curb-ramp register was at least as informative as new street imagery, whereas for buildings in poorly mapped areas, photographs added information that the map lacked. The floor-count gradient suggests a simple screening variable: the distance from each building to the nearest mapped value, which can be computed from existing data before any image is collected. Areas where this distance is large and the target is visually legible are the natural candidates for new imagery. \citet{ki2026weather} showed that filtering imagery for consistent weather reduces spatial coverage. For floor counts, our results suggest that retaining coverage may be especially useful where the target is visible and the nearest labelled building is distant.

Source value also depends on how often an attribute changes. Imagery can capture visible changes in urban conditions, while stable attributes may already be recorded in administrative data. For features that do change, image date matters, and our Project Sidewalk check showed that replacing current panoramas with source-era views can change many curb-ramp predictions (Supplementary Section B).

For users of derived maps, the relevant question is which source produced each value. When a model receives both the image and existing records, its predictions in data-sparse areas may depend mainly on the photograph, whereas in well-recorded areas they may follow nearby labels or administrative records. Capture dates and record dates also matter for urban assets that change. These differences are lost when every output is described simply as image-derived.

\subsection{Implications for model evaluation and dataset release}

For researchers who evaluate multimodal urban models, the source-conflict results show that reported performance can depend on what is supplied with the image. Descriptions, points of interest, and reference values raised accuracy in our experiments without any contribution from the photograph. Reporting image-only and context-only results alongside the combined condition would show how much of the performance comes from the image. A comparison with the strongest available non-image source would show whether the image is worth collecting at all. For trained models, spatially blocked validation serves a similar purpose \citep{roberts2017crossvalidation}.

For producers of derived datasets, the OpenFACADES case shows that the annotation process can itself introduce dependence on existing data. The annotation process is one of the sources of bias identified for language technologies \citep{hovy2021bias}. The released floor labels were generated from both imagery and OpenStreetMap attributes, and their high accuracy for tall buildings most likely reflects the record rather than the image. Such labels remain useful, provided their inputs are known. Dataset documentation should state which inputs the annotating model received \citep{gebru2021datasheets}. Later studies can then avoid treating the labels as independent image-only ground truth, or evaluating image models against values that partly came from the same map.

\section{Limitations}
\label{sec:limitations}

The public resources differ in geography, sampling, task definition, and image date. Each attribute comes from one resource, and three of the five resources cover a single city, so differences between attributes are partly differences between these settings. Project Sidewalk presents the clearest temporal mismatch because many retrieved panoramas are newer than the crowd labels. A rerun with archived source-era panoramas changed many individual predictions, but the register remained more accurate than both sets of image results (Supplementary Section B). OpenFACADES images and the additional OpenStreetMap attributes retrieved in 2026 also come from different years, which can affect buildings that changed between image capture and record retrieval. The CityLens population condition includes satellite imagery, so its image contribution cannot be attributed to street views alone. The US house-price reference is assigned at ZIP-code level, so the comparison cannot assess price variation within ZIP codes.

The study evaluates selected public datasets and three lightweight general-purpose VLMs, and the results describe the information settings represented by those resources. A specialised detector, a larger VLM, a more complete register, or another image viewpoint could change the balance between sources. The literature review covers only studies with accessible full text and was coded by one person.

\section{Conclusion}

Street-view imagery adds the most information when an urban attribute is visible and existing records are sparse. This value depends on place as well as on the attribute. It changes across a city with the density of existing labels, and it can be masked when a model also receives records that it tends to follow. Evaluating imagery against the data already available for the same place gives a clearer basis for deciding where new images are worth collecting. Documenting which sources enter each prediction or annotation shows how the resulting maps and datasets should be used.

%% file: sec/9_app.tex
\appendix

\section{Literature Review and Study Selection}
\label{app:audit}

We searched OpenAlex for studies published since 2020 that used street-level imagery to predict objective building attributes. Screening of titles, abstracts, and full texts reduced 2,041 records to 33 unique open-access studies (Figure S1a).

No retained study reported a predictor based only on location or neighbouring labels (Figure S1b). Other reporting practices were uneven: 17 studies reported any baseline, and four each reported spatial cross-validation or uncertainty. One coder completed the review. The companion archive provides the article table, coding decisions, codebook, and supporting evidence.

\begin{figure}[H]
  \centering
  \includegraphics[width=\linewidth]{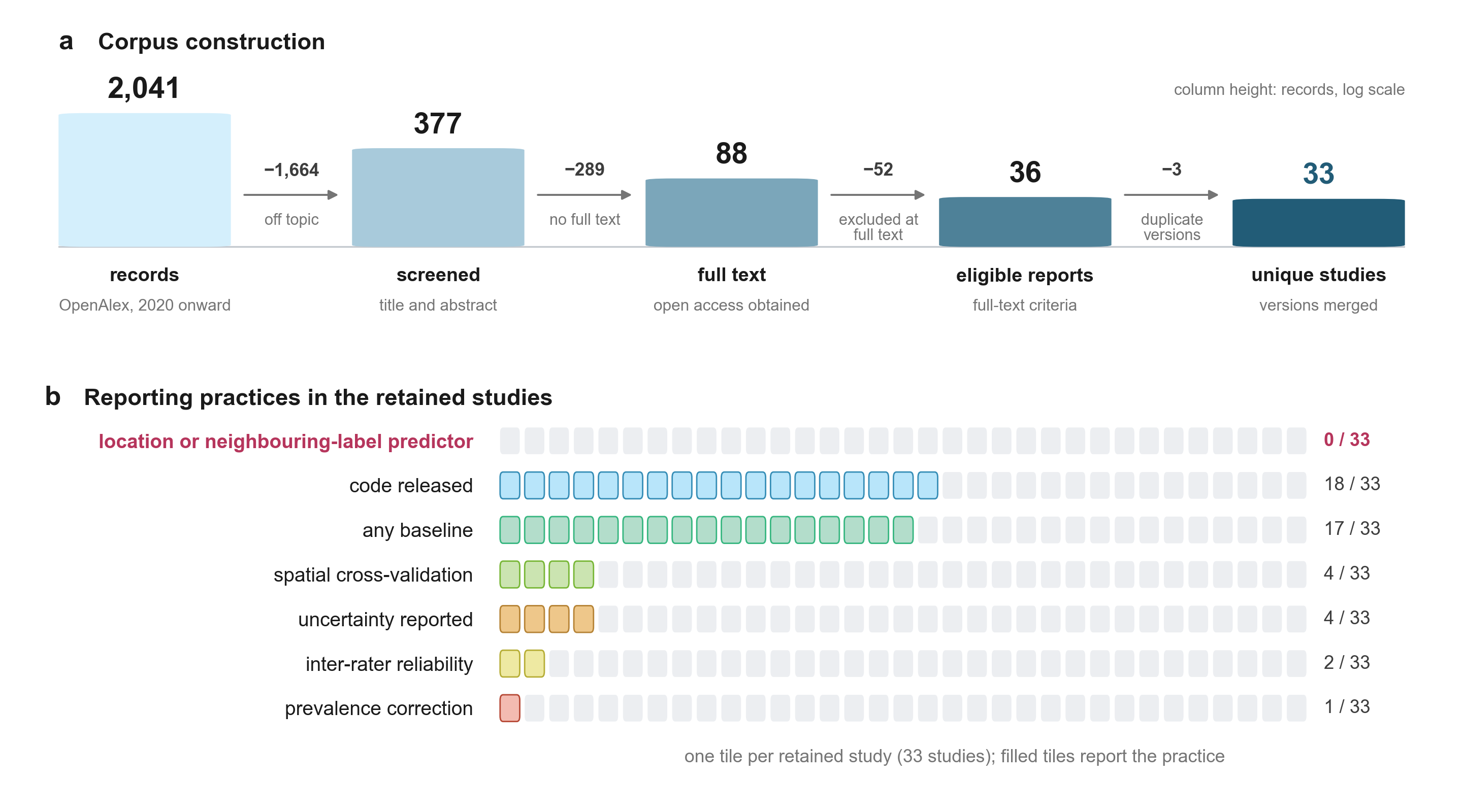}
  \caption{Selection and reporting in the reviewed street-view
  building-attribute literature. \textbf{(a)}~Records followed through title,
  abstract, and full-text screening; column height is on a log scale.
  \textbf{(b)}~Reporting practices among the 33 retained studies, one tile per
  study; filled tiles mark studies that report the practice.}
  \label{fig:audit-supp}
\end{figure}

\section{Additional Design Details and Sensitivity Analyses}
\label{app:temporal}

\paragraph{Non-image source selection.}
The non-image alternative differs by task because each resource represents a different urban data setting (Table S1). Road damage uses nearby observations from the same survey, curb ramps use an independent municipal register, and the remaining tasks use labels, OpenStreetMap (OSM) tags, or statistics from nearby units. The main comparison uses the strongest tested source for each task. For CityLens, this source is the median score of the five nearest other areas; a VLM that receives the same statistics without an image is reported in Supplementary Section J (Table S12). Results for all candidate sources and spatial settings are in the companion archive.

\begin{table}[H]
  \caption{Existing urban sources considered for each task.}
  \label{tab:comparator-provenance}
  \centering
  \scriptsize
  \begin{tabular}{llll}
    \toprule
    {\tableheader Attribute} & {\tableheader Sources considered} &
    {\tableheader Main comparison source} &
    {\tableheader Urban setting} \\
    \midrule
    Road damage & prevalence, nearby frames & nearby frames & gaps in a dense survey \\
    \rowgray Curb ramp & prevalence, nearby labels, SDOT & SDOT register & municipal inventory \\
    Floor count & prevalence, nearby OSM & nearby OSM tags & incomplete building inventory \\
    \rowgray Building type & prevalence, nearby OSM & modal OSM class & local class distribution \\
    Building function & prevalence, nearby labels & nearby labels & sampled local labels \\
    \rowgray Population & prevalence, official neighbours & official neighbours & area statistics \\
    House price & prevalence, official neighbours & official neighbours & tract statistics \\
    \bottomrule
  \end{tabular}
\end{table}

\begin{figure}[H]
  \centering
  \includegraphics[width=\linewidth]{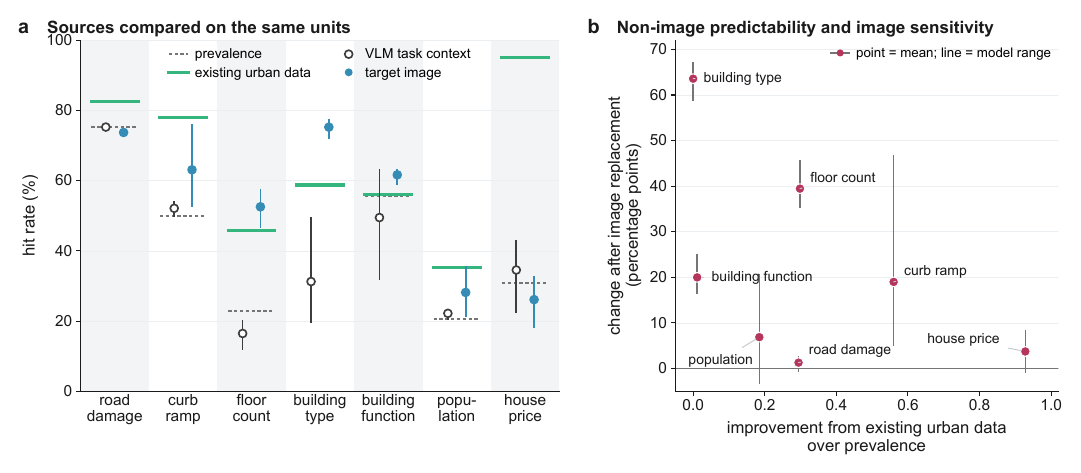}
  \caption{How prediction sources vary across tasks. \textbf{(a)}~Performance
  from prevalence, existing urban data, task context, and target images.
  \textbf{(b)}~Relationship between non-image predictability and sensitivity
  to image replacement. Values are interpreted within each task because the
  metrics and class structures differ. Both panels use hit rates; for road
  damage this is the mean per-class accuracy, under which always predicting
  no damage scores 75.3\%.}
  \label{fig:evidence-map-supp}
\end{figure}

\paragraph{SVRDD exclusion radius.}
We repeated the road-damage comparison with progressively wider gaps between the target frame and its labelled neighbours. Balanced accuracy declined from 72.3\% at 50 m to 64.3\% at 400 m, and macro accuracy from 82.6\% to 76.5\% (Table S2). The 400 m macro-accuracy interval includes the prevalence benchmark of 75.3\%, whereas balanced accuracy stayed well above the corresponding benchmark of 50\%. The image-only models (50.4\% to 54.7\% balanced accuracy, Table S3) stayed below the spatial predictor at every radius.

\begin{table}[H]
  \caption{SVRDD buffered spatial predictor at four exclusion radii. Both
  metrics average over six damage classes; brackets give bootstrap 95\%
  intervals.}
  \label{tab:svrdd-buffer}
  \centering
  \begin{tabular}{lcccc}
    \toprule
    {\tableheader Exclusion radius} & 50\,m & 100\,m & 200\,m & 400\,m \\
    \midrule
    Balanced accuracy & 72.3 [69.1, 75.3] & 69.4 [66.3, 72.5] &
    66.3 [63.0, 69.5] & 64.3 [61.2, 67.4] \\
    \rowgray Macro accuracy & 82.6 [80.6, 84.6] & 80.3 [78.3, 82.3] &
    77.8 [75.7, 80.0] & 76.5 [74.4, 78.5] \\
    \bottomrule
  \end{tabular}
\end{table}

The nearby-observation predictor also led under metrics that give positive damage classes more weight (Table S3). Unlike the prevalence benchmark, it identified positive damage, with a positive recall of 54.9\% against 4.8\% to 23.7\% for the images.

\begin{table}[H]
  \caption{SVRDD results under class-sensitive metrics at the 50\,m exclusion
  radius. Values are percentages averaged over the six damage classes.}
  \label{tab:svrdd-balanced}
  \centering
  \small
  \begin{tabular}{lrrrr}
    \toprule
    {\tableheader Predictor} & {\tableheader Macro acc.} &
    {\tableheader Macro-F1} & {\tableheader Balanced acc.} &
    {\tableheader Positive recall} \\
    \midrule
    Prevalence benchmark & 75.3 & 0.0 & 50.0 & 0.0 \\
    \rowgray Buffered spatial & \best{82.6} & \best{59.6} & \best{72.3} & \best{54.9} \\
    GPT image only & 74.0 & 7.9 & 50.4 & 4.8 \\
    \rowgray Gemini image only & 73.9 & 27.5 & 54.7 & 23.7 \\
    Qwen image only & 73.3 & 23.6 & 52.6 & 19.7 \\
    \bottomrule
  \end{tabular}
\end{table}

\paragraph{Project Sidewalk timing.}
Each evaluated corner corresponds to a cluster of crowd labels, and the public cluster exports give the mean capture date of the source images and the mean label timestamp. The panoramas we retrieved were generally newer than these dates, so we also recovered archived source-era panoramas from Project Sidewalk's public backups. For each cluster, we selected the panorama referenced most often by its labels and centred the view on the labelled location. These images contained no label overlay. A matched subset of 100 clusters was rerun with the same prompts and model settings as the main comparison.

Image date changed about 30\% of individual predictions for GPT-4o-mini and Gemini, and its effect on balanced accuracy differed in direction across models (Table S4). The municipal register was more accurate than both sets of image results. Image date can therefore affect predictions for street features that change over time.

\begin{table}[H]
  \caption{Matched Project Sidewalk temporal validation on 100 clusters
  (50 per class). The interval is a paired stratified-bootstrap 95\%
  interval for the source-era-minus-current balanced-accuracy difference.}
  \label{tab:pssw-time}
  \centering
  \begin{tabular}{lcccc}
    \toprule
    {\tableheader Model} & {\tableheader Current} &
    {\tableheader Source-era} & {\tableheader Difference} &
    {\tableheader Changed} \\
    \midrule
    GPT-4o-mini & 80.0\% & 74.0\% &
    $-6.0$ [$-17.0$, $+5.0$] & 30.0\% \\
    Gemini 2.5 Flash Lite & 63.0\% & 68.0\% &
    $+5.0$ [$-5.0$, $+16.0$] & 31.0\% \\
    Qwen3.5 Flash & 50.0\% & 54.0\% &
    $+4.0$ [$+1.0$, $+8.0$] & 4.0\% \\
    \bottomrule
  \end{tabular}
\end{table}

\paragraph{Image replacement.}
Donor images are drawn from other units under task-specific constraints, and the target's task context is kept (Table S5). Road donors come from another district, curb-ramp and building-type donors belong to a different reference class, and CityLens donors come from another city. Building-function donors are sampled independently, so some share the target's class.

\begin{table}[H]
  \caption{Reference-label separation between target and donor images.}
  \label{tab:swap-separation}
  \centering
  \scriptsize
  \begin{tabular}{ll}
    \toprule
    {\tableheader Task} & {\tableheader Donor separation} \\
    \midrule
    Road damage & another district, with at least one different damage label \\
    \rowgray Curb ramp & opposite reference class \\
    Floor count & another building, usually differing by more than one floor \\
    \rowgray Building type & different reference class \\
    Building function & independently sampled building \\
    \rowgray Population & area in another city \\
    House price & tract in another city \\
    \bottomrule
  \end{tabular}
\end{table}

\paragraph{Building-type class balance.}
The building-type sample is stratified by floor band, contains six reference classes, and is dominated by apartments. Macro-F1 and balanced accuracy weight each observed class equally, and predictions into other offered classes count as errors (Table S6).

\begin{table}[H]
  \caption{Building-type results on the floor-stratified sample. Values are
  percentages.}
  \label{tab:btype-balanced}
  \centering
  \small
  \begin{tabular}{lrrr}
    \toprule
    {\tableheader Predictor} & {\tableheader Accuracy} &
    {\tableheader Macro-F1} & {\tableheader Balanced accuracy} \\
    \midrule
    Modal class & 58.8 & 12.3 & 16.7 \\
    \rowgray Spatial predictor & 27.8 & 18.7 & 25.2 \\
    GPT image only & 72.2 & 64.1 & 70.8 \\
    \rowgray Gemini image only & 77.3 & 74.2 & 76.1 \\
    Qwen image only & 76.3 & 64.3 & 63.1 \\
    \bottomrule
  \end{tabular}
\end{table}

\section{Floor-Count Data and Spatial Variation}
\label{app:floorcount}

\paragraph{Sample and sites.}
We retained the OSM-derived floor values and building identifiers from the OpenFACADES release and retrieved additional OSM attributes and neighbouring floor tags. The seven metropolitan areas differ in both reference density and building height, and the evidence on tall buildings comes mainly from Berlin, San Francisco, Manila, Amsterdam, and Houston.

\paragraph{Image-to-building pairing.}
The release links each perspective crop to a building ID. We computed the distance from the recorded camera position to the building's OpenStreetMap footprint and used it in the adjusted analysis. This distance increased with building height. Very close views of tall buildings often lost the roofline, while distant views lost facade detail. The photographs were taken between 2012 and 2024; additional target attributes and neighbouring OSM tags were retrieved in 2026. Floor-count references were retained from the release, and accuracy showed no monotonic relationship with capture year.

\paragraph{Reference inspection.}
Without viewing model predictions, we inspected images sampled from each height group together with their OSM records. Low-rise facades were usually countable. Most sampled high-rise images omitted the roofline, were taken at a steep angle, or contained substantial occlusion. For buildings with both height and level tags, the implied storey heights were generally plausible.

\paragraph{Exact floor-count prediction.}
When a prediction had to match the reference floor count exactly, images outperformed the spatial predictor for low-rise buildings and underperformed it for tall buildings, in line with the visibility review.

\paragraph{Variation across spatial settings.}
The pooled image advantage stayed positive across buffer widths and neighbour counts, after omitting each metropolitan area, and after moderate error was introduced into the reference pool. The advantage remained larger where nearby mapped values were sparse.

\paragraph{Adjusted reference-density gradient.}
For each building $i$, $D_i$ is the mean image-only correctness across available models minus the correctness of the spatial predictor. We estimate
\begin{equation}
  D_i = \beta_0 + \beta_1\log_2 d_i
  + \alpha_{\mathrm{metro}(i)} + \gamma_{\mathrm{height}(i)}
  + \beta_2\log_2 c_i + \beta_3(y_i-\widetilde{y}) + \varepsilon_i,
  \label{eq:density-gradient}
\end{equation}
where $d_i$ is the distance to the nearest labelled building, $c_i$ is the camera-to-building distance (bounded below at 1 m before taking the logarithm), and $y_i$ is the capture year. The median capture year, $\widetilde{y}$, is 2022. Metropolitan area and building-height group enter as categorical terms, with Amsterdam and 15 or more floors as the reference categories. We used 4,000 bootstrap resamples within metropolitan-area-by-height cells. Four low-rise buildings in Amsterdam are omitted because no camera position or capture date could be retrieved from Mapillary for their images, which leaves 444 of the 448 buildings.

\begin{table}[H]
  \caption{Full estimates for the adjusted image-advantage model in
  Equation~\ref{eq:density-gradient}. Coefficients are percentage points.
  Bootstrap standard errors and percentile intervals use 4,000 resamples
  within metropolitan-area-by-height cells. Reference categories are
  Amsterdam and buildings with 15 or more floors.}
  \label{tab:density-model}
  \centering
  \small
  \begin{tabular}{lrrr}
    \toprule
    {\tableheader Term} & {\tableheader Estimate} &
    {\tableheader Bootstrap SE} & {\tableheader 95\% interval} \\
    \midrule
    Intercept & $-96.07$ & $27.31$ & $[-148.65,-42.20]$ \\
    $\log_2$ nearest-label distance (m) & $+5.74$ & $2.18$ & $[+1.42,+10.03]$ \\
    $\log_2$ camera-to-building distance (m) & $+9.12$ & $4.04$ & $[+1.19,+16.89]$ \\
    Capture year minus 2022 & $+0.94$ & $2.47$ & $[-3.95,+5.72]$ \\
    Height: 1--4 floors & $+43.92$ & $6.55$ & $[+30.31,+56.31]$ \\
    Height: 5--14 floors & $+19.81$ & $5.97$ & $[+7.98,+31.64]$ \\
    Metropolitan area: Berlin & $+2.53$ & $17.19$ & $[-30.48,+37.32]$ \\
    Metropolitan area: Helsinki & $-1.07$ & $14.91$ & $[-28.58,+29.37]$ \\
    Metropolitan area: Houston & $+18.24$ & $11.05$ & $[-3.67,+39.47]$ \\
    Metropolitan area: Manila & $+11.85$ & $19.29$ & $[-24.49,+50.31]$ \\
    Metropolitan area: San Francisco & $+13.62$ & $16.42$ & $[-17.73,+46.38]$ \\
    Metropolitan area: Washington & $-17.01$ & $12.87$ & $[-41.34,+9.04]$ \\
    \midrule
    Buildings & \multicolumn{3}{r}{444} \\
    Model--building observations & \multicolumn{3}{r}{1,328} \\
    $R^2$ & \multicolumn{3}{r}{0.143} \\
    \bottomrule
  \end{tabular}
\end{table}

Nearest-label distances in the regression sample ranged from 50 to 2,366 m. The fitted image advantage increased with distance, and re-estimating the model without each metropolitan area gave slopes of 3.45 to 7.74 percentage points per doubling, all positive.

\section{Supervised Replication}
\label{app:supervised}

Many street-view studies train a prediction head on fixed visual features rather than prompting a VLM. To check whether the floor-count pattern depends on the VLM setting, we repeated the comparison with features from an ImageNet-pretrained ResNet-50 network and a linear classifier. The classifier and the spatial predictor were evaluated under random and geographically blocked splits, the latter being recommended when observations are spatially correlated \citep{roberts2017crossvalidation}. Blocking reduced performance for both, but the spatial predictor declined more, so the image advantage widened from about 12 to 21 percentage points. Blocked splits leave fewer labelled buildings near each test building, which matches the conditions under which the main analysis found imagery most useful.

\section{Additional Attribute Comparison}
\label{app:constructs}

This comparison checks whether image gains over coordinates arise only for attributes that can be seen. Four attributes were scored from the same OpenFACADES images. Floor count, building type, and construction decade have potential facade cues. Distance to the nearest hospital is spatially patterned but cannot be seen in the photograph, and it serves as a negative control (Table S8 and Figure S3). The reference search included OSM hospital and clinic tags, although the prompt asked for hospitals; this mismatch limits the interpretation of absolute accuracy on this task. Images outperformed coordinates by 30 to 37 percentage points for the three facade attributes, but not for hospital distance ($-$2.0 points).

\begin{table}[H]
  \caption{Image and coordinate accuracy for three facade attributes and
  distance to the nearest hospital. $n$ counts paired model--building
  observations across the three models. A construction-decade answer, a
  year rounded to the decade, counts as correct when it lies within 15 years
  of the OpenStreetMap start date; for a start date of 1968, answers of
  1960, 1970, and 1980 are correct.}
  \label{tab:constructs}
  \centering
  \small
  \begin{tabular}{lcrrr}
    \toprule
    {\tableheader Attribute} & {\tableheader $n$} & {\tableheader Image} &
    {\tableheader Coordinates} & {\tableheader Difference} \\
    \midrule
    Floors, within 1 & 450 & 50.7\% & 17.3\% & \best{$+33.3$} \\
    \rowgray Building type & 291 & 76.6\% & 46.4\% & \best{$+30.2$} \\
    Construction decade & 93 & 71.0\% & 34.4\% & \best{$+36.6$} \\
    \rowgray Distance to hospital & 450 & 20.4\% & 22.4\% & $-2.0$ \\
    \bottomrule
  \end{tabular}
\end{table}

\begin{figure}[H]
  \centering
  \includegraphics[width=0.85\linewidth]{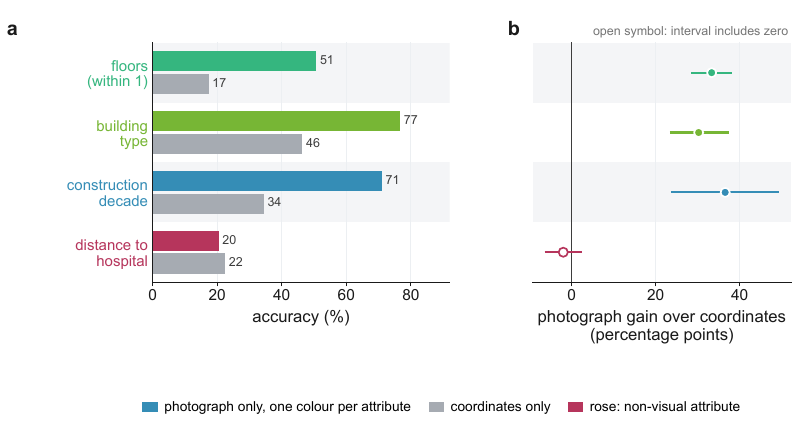}
  \caption{Image and coordinate performance for three facade attributes and
  distance to the nearest hospital. Bars and points are coloured by
  attribute, with rose marking the non-visual attribute;
  horizontal lines show paired uncertainty intervals.}
  \label{fig:constructs}
\end{figure}

Given coordinates alone, GPT-4o-mini and Gemini returned almost the same hospital-distance band for most buildings. Image-based responses for the facade attributes were more varied and closer to the reference distributions. This pattern suggests that the facade predictions draw on visual information rather than repeating the dominant class.

\section{BuildingSense Class-Balanced Results}
\label{app:bsense-classes}

The BuildingSense sample is dominated by residential buildings and contains few examples of several classes. We therefore report macro-F1 and balanced accuracy alongside overall accuracy. Under both class-balanced measures, all three image models exceeded the nearby-label predictor (Table S9). Table S10 gives the paired contrasts used in Section 5.3: the change in context-only accuracy when nearby points of interest are added to the description, the change when the address is replaced, and the image gain for buildings whose description names no nearby place. Per-class results are provided in the companion archive.

\begin{table}[H]
  \caption{BuildingSense results under overall and class-balanced metrics.
  Macro metrics average over the 21 reference classes present; invalid
  predictions remain errors.}
  \label{tab:bsense-balanced}
  \centering
  \begin{tabular}{lccc}
    \toprule
    {\tableheader Predictor} & {\tableheader Accuracy} &
    {\tableheader Macro-F1} & {\tableheader Balanced accuracy} \\
    \midrule
    Spatial (50\,m) & 56.0\% & 6.5\% & 8.4\% \\
    \rowgray GPT image only & 63.0\% & 23.9\% & 27.6\% \\
    Gemini image only & 59.0\% & 24.2\% & 24.0\% \\
    \rowgray Qwen image only & 63.0\% & 25.4\% & 31.1\% \\
    \bottomrule
  \end{tabular}
\end{table}

\begin{table}[H]
  \caption{Paired BuildingSense source contrasts in percentage points, with
  uncertainty intervals.}
  \label{tab:bsense-ablation-ci}
  \centering
  \scriptsize
  \begin{tabular}{lccc}
    \toprule
    {\tableheader Model} & {\tableheader Nearby place added} &
    {\tableheader Address shuffle} & {\tableheader Image added, no nearby place} \\
    \midrule
    GPT & $+23.1\ [10.8,35.4]$ & $+4.5\ [-1.0,10.5]$ & $+14.1\ [8.1,20.7]$ \\
    \rowgray Gemini & $+21.5\ [7.7,35.4]$ & $-1.0\ [-8.0,6.5]$ & $+44.4\ [35.6,53.3]$ \\
    Qwen & $+18.5\ [6.2,30.8]$ & $+3.0\ [-2.5,8.5]$ & $+8.9\ [1.5,16.3]$ \\
    \bottomrule
  \end{tabular}
\end{table}

\section{Height Pattern in Annotation Errors}
\label{app:power}

The error curves in Figure 6b use the same geographically matched buildings for the image-only reruns and the released annotations. The floor-band matrices in Figure 6c,d use 1,340 image-only predictions and 4,339 released annotations, respectively, rather than a matched sample. Image-only error increased with reference floor count. Released-annotation error stayed low up to 30 floors, and the share of released labels in the correct floor band rose with height (Figure 6b--d). Model predictions were averaged within each building so that each building counts once in the height relationship.

This reversal is consistent with greater reliance on the supplied floor record when visual evidence becomes less informative. In the record-conflict experiment, models also followed supplied values most strongly for tall buildings.

\section{Reference Influence in OpenFACADES Annotations}
\label{app:forensics}

\begin{figure}[H]
  \centering
  \includegraphics[width=\linewidth]{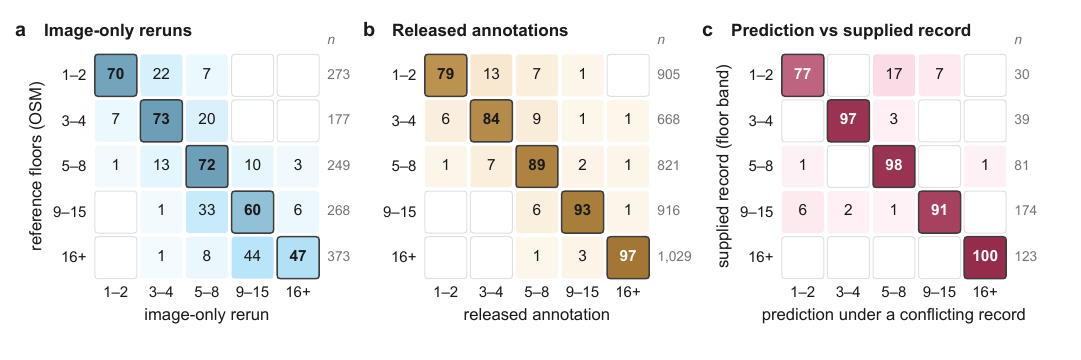}
  \caption{Floor-band confusion for \textbf{(a)} image-only reruns,
  \textbf{(b)} released OpenFACADES annotations, and \textbf{(c)} predictions
  scored against a supplied conflicting record. Cells give row percentages,
  normalised within each band of the reference floors in (a) and (b) and of
  the supplied record in (c); $n$ is the number of cases in each row.}
  \label{fig:confusion}
\end{figure}

\begin{figure}[H]
  \centering
  \includegraphics[width=\linewidth]{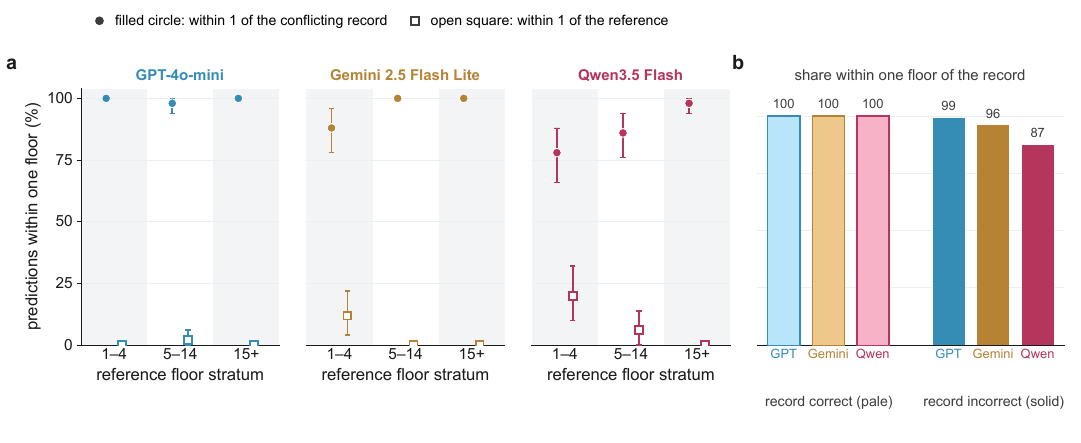}
  \caption{Floor-count predictions under record conflict.
  \textbf{(a)}~Agreement with the conflicting record and the independent
  reference by building height and model. \textbf{(b)}~Agreement when the
  supplied record is correct or incorrect.}
  \label{fig:causalfloors-supp}
\end{figure}

The released workflow attaches OpenStreetMap attributes to each image before a commercial VLM writes the caption. Some captions explicitly attribute a value to OpenStreetMap, and the released floor labels show the height pattern described in Supplementary Section G. The record-conflict experiment shows that supplying the same type of reference value can substantially change a new prediction. When the supplied record was wrong, 87\% to 99\% of predictions still fell within one floor of it (Figure S5b).

\section{Prompt Design}
\label{app:prompts}

All prompts request a fixed-format answer and use temperature zero. Full prompt templates and answer schemas are in the companion archive.

For floor count, the basic instruction asks the model to inspect the building. Location conditions add coordinates, and record conditions state a floor value in the style of an address record. Building-type prompts follow the same structure and ask for one of the mapped OpenStreetMap classes.

The road-damage prompt asks which damage classes appear on the road surface, and its context gives the Beijing district and coordinates. The conflicting record describes a recently resurfaced segment with no reported damage. The curb-ramp prompt asks whether the corner has a ramp, and its record conditions use the wording and proximity information of an asset register.

BuildingSense uses the released image, text, and combined conditions. Paired text runs remove nearby points of interest, keep selected fields, or replace the address. Qwen comparisons use the same reasoning configuration within each pair.

CityLens uses the released task wording. Population runs include ten street views and one satellite image per area; house-price runs use ten street views per tract. Image replacements exchange the full bundle while retaining the target prompt. Location-only runs replace the images with the coordinates of the area centroid, and statistics-input runs add the official scores of the five nearest areas to the location-only prompt. Cross-city image replacements retain the target prompt, including its city name; a separate image-only condition removes the city name. Full model settings are in the companion archive.

\section{Additional Paired Results}
\label{app:matrices}

Full results for all model and source conditions, including the class-level and task-specific comparisons omitted from the main text, are in the companion archive. Table S11 reports the paired image-replacement results shown in Figure 5a. Table S12 compares the released CityLens image prompt with the statistics-input condition, in which the VLM receives the location-only prompt together with the scores of the five nearest areas. The statistics input gave lower or similar MAE for every model and task.

\paragraph{Invalid and incomplete model responses.}
Most conditions produced a valid answer for nearly every unit. For the CityLens house-price task, only 10 of the 120 tracts had valid Gemini answers in both the original and image-replacement conditions, and Figure 5a marks this estimate with an open symbol. Invalid BuildingSense taxonomy outputs were counted as errors. Raw responses and condition-level counts are in the companion archive.

\begin{table}[H]
  \caption{Paired image-replacement effects in percentage points. Positive
  values indicate lower accuracy after replacement. Road damage is scored
  with balanced accuracy and the CityLens tasks with the share of
  predictions within one point of the reference. Brackets give uncertainty
  intervals and parentheses give paired sample sizes.}
  \label{tab:swap-intervals}
  \centering
  \scriptsize
  \begin{tabular}{lccc}
    \toprule
    {\tableheader Task} & {\tableheader GPT} & {\tableheader Gemini} &
    {\tableheader Qwen} \\
    \midrule
    Road damage & $+0.8\ [-0.7,2.4]\ (300)$ & $+4.6\ [1.9,7.3]\ (300)$ & $+4.5\ [1.8,7.3]\ (300)$ \\
    \rowgray Curb ramp & $+46.7\ [39.7,53.7]\ (300)$ & $+5.0\ [1.3,8.7]\ (300)$ & $+5.4\ [2.7,8.4]\ (299)$ \\
    Floor count & $+35.3\ [26.7,44.0]\ (150)$ & $+45.6\ [36.2,55.0]\ (149)$ & $+37.4\ [28.6,46.3]\ (147)$ \\
    \rowgray Building type & $+58.8\ [46.4,70.1]\ (97)$ & $+67.0\ [55.7,78.4]\ (97)$ & $+64.9\ [53.6,75.3]\ (97)$ \\
    Building function & $+16.5\ [10.0,23.0]\ (200)$ & $+18.5\ [11.0,26.0]\ (200)$ & $+25.0\ [19.0,31.5]\ (200)$ \\
    \rowgray Population & $-3.3\ [-10.0,3.3]\ (150)$ & $+3.4\ [-6.9,13.8]\ (116)$ & $+20.5\ [10.3,30.8]\ (146)$ \\
    House price & $-0.8\ [-9.2,7.5]\ (120)$ & $+20.0\ [0.0,50.0]\ (10)$ & $+8.3\ [-0.8,17.5]\ (120)$ \\
    \bottomrule
  \end{tabular}
\end{table}

\begin{table}[H]
  \caption{Released CityLens image prompt compared with the statistics-input
  condition, in which the VLM receives the location-only prompt and the
  scores of the five nearest areas but no image. MAE uses the released score
  scale and is computed on paired areas. The difference is the
  statistics-input MAE minus the image-prompt MAE, so negative values favour
  the statistics input. Brackets give uncertainty intervals and parentheses
  give paired sample sizes.}
  \label{tab:statistics-input}
  \centering
  \scriptsize
  \begin{tabular}{llccc}
    \toprule
    {\tableheader Task} & {\tableheader Model} & {\tableheader Image prompt} &
    {\tableheader Statistics input} & {\tableheader Difference} \\
    \midrule
    Population & GPT & 2.76 & 1.88 & $-0.87\ [-1.13,-0.63]\ (150)$ \\
    \rowgray & Gemini & 2.18 & 1.90 & $-0.27\ [-0.64,0.07]\ (123)$ \\
    & Qwen & 1.97 & 1.98 & $+0.01\ [-0.34,0.37]\ (148)$ \\
    \rowgray House price & GPT & 2.94 & 0.25 & $-2.69\ [-3.14,-2.28]\ (120)$ \\
    & Gemini & 4.38 & 0.41 & $-3.97\ [-5.94,-2.00]\ (12)$ \\
    \rowgray & Qwen & 1.85 & 0.24 & $-1.61\ [-1.84,-1.38]\ (120)$ \\
    \bottomrule
  \end{tabular}
\end{table}

%% file: main.bbl
\begin{thebibliography}{}

\bibitem[Biljecki et~al., 2023a]{biljecki2023buildingquality}
Biljecki, F., Chow, Y.~S., and Lee, K. (2023a).
\newblock Quality of crowdsourced geospatial building information: A global
  assessment of {OpenStreetMap} attributes.
\newblock {\em Building and Environment}, 237:110295.

\bibitem[Biljecki and Ito, 2021]{biljecki2021svi}
Biljecki, F. and Ito, K. (2021).
\newblock Street view imagery in urban analytics and {GIS}: A review.
\newblock {\em Landscape and Urban Planning}, 215:104217.

\bibitem[Biljecki et~al., 2023b]{biljecki2023sensitivity}
Biljecki, F., Zhao, T., Liang, X., and Hou, Y. (2023b).
\newblock Sensitivity of measuring the urban form and greenery using
  street-level imagery: A comparative study of approaches and visual
  perspectives.
\newblock {\em International Journal of Applied Earth Observation and
  Geoinformation}, 122:103385.

\bibitem[Chen et~al., 2024]{chen2024mmstar}
Chen, L., Li, J., Dong, X., Zhang, P., Zang, Y., Chen, Z., Duan, H., Wang, J.,
  Qiao, Y., Lin, D., and Zhao, F. (2024).
\newblock Are we on the right way for evaluating large vision-language models?
\newblock In {\em Advances in Neural Information Processing Systems},
  volume~37, pages 27056--27087. Curran Associates, Inc.

\bibitem[Crooks and Chen, 2024]{crooks2024information}
Crooks, A. and Chen, Q. (2024).
\newblock Exploring the new frontier of information extraction through large
  language models in urban analytics.
\newblock {\em Environment and Planning B: Urban Analytics and City Science},
  51(3):565--569.

\bibitem[Dai et~al., 2025]{dai2025busstops}
Dai, Y., Liu, L., Wang, K., Li, M., and Yan, X. (2025).
\newblock Using computer vision and street view images to assess bus stop
  amenities.
\newblock {\em Computers, Environment and Urban Systems}, 117:102254.

\bibitem[Fan et~al., 2025]{fan2025coverage}
Fan, Z., Feng, C.-C., and Biljecki, F. (2025).
\newblock Coverage and bias of street view imagery in mapping the urban
  environment.
\newblock {\em Computers, Environment and Urban Systems}, 117:102253.

\bibitem[Gebru et~al., 2017]{gebru2017demographics}
Gebru, T., Krause, J., Wang, Y., Chen, D., Deng, J., Aiden, E.~L., and Fei-Fei,
  L. (2017).
\newblock Using deep learning and {Google Street View} to estimate the
  demographic makeup of neighborhoods across the {United States}.
\newblock {\em Proceedings of the National Academy of Sciences},
  114(50):13108--13113.

\bibitem[Gebru et~al., 2021]{gebru2021datasheets}
Gebru, T., Morgenstern, J., Vecchione, B., Vaughan, J.~W., Wallach, H.,
  Daum{\'e}, III, H., and Crawford, K. (2021).
\newblock Datasheets for datasets.
\newblock {\em Communications of the ACM}, 64(12):86--92.

\bibitem[Geirhos et~al., 2020]{geirhos2020shortcut}
Geirhos, R., Jacobsen, J.-H., Michaelis, C., Zemel, R., Brendel, W., Bethge,
  M., and Wichmann, F.~A. (2020).
\newblock Shortcut learning in deep neural networks.
\newblock {\em Nature Machine Intelligence}, 2(11):665--673.

\bibitem[Goyal et~al., 2017]{goyal2017vqa}
Goyal, Y., Khot, T., Summers-Stay, D., Batra, D., and Parikh, D. (2017).
\newblock Making the {V} in {VQA} matter: Elevating the role of image
  understanding in visual question answering.
\newblock In {\em 2017 IEEE Conference on Computer Vision and Pattern
  Recognition}, pages 6325--6334. IEEE.

\bibitem[Hou and Biljecki, 2022]{hou2022sviquality}
Hou, Y. and Biljecki, F. (2022).
\newblock A comprehensive framework for evaluating the quality of street view
  imagery.
\newblock {\em International Journal of Applied Earth Observation and
  Geoinformation}, 115:103094.

\bibitem[Hovy and Prabhumoye, 2021]{hovy2021bias}
Hovy, D. and Prabhumoye, S. (2021).
\newblock Five sources of bias in natural language processing.
\newblock {\em Language and Linguistics Compass}, 15(8):e12432.

\bibitem[Jang and Kim, 2025]{jang2025auditing}
Jang, K.~M. and Kim, J. (2025).
\newblock Multimodal large language models as built environment auditing tools.
\newblock {\em The Professional Geographer}, 77(1):84--90.

\bibitem[Ki et~al., 2025]{ki2025walkability}
Ki, D., Lee, H., Park, K., Ha, J., and Lee, S. (2025).
\newblock Measuring nuanced walkability: Leveraging {ChatGPT}'s vision
  reasoning with multisource spatial data.
\newblock {\em Computers, Environment and Urban Systems}, 121:102319.

\bibitem[Ki et~al., 2026]{ki2026weather}
Ki, D., Lee, S., Han, C., Kim, Y., Koo, B.~W., and Hwang, U. (2026).
\newblock Does weather matter? {Examining} measurement bias in street view
  image-based urban perception assessments.
\newblock {\em Computers, Environment and Urban Systems}, 127:102421.

\bibitem[Kim et~al., 2026]{kim2026geographic}
Kim, H., Li, X., Quintana, M., Biljecki, F., and Lee, S. (2026).
\newblock Geographic and perceptual bias in multimodal {LLMs}: Evidence from a
  global dataset spanning more than 200 cities.
\newblock {\em npj Urban Sustainability}.
\newblock \url{https://doi.org/10.1038/s42949-026-00466-2}.

\bibitem[Kim et~al., 2021]{kim2021measurement}
Kim, J.~H., Lee, S., Hipp, J.~R., and Ki, D. (2021).
\newblock Decoding urban landscapes: {Google Street View} and measurement
  sensitivity.
\newblock {\em Computers, Environment and Urban Systems}, 88:101626.

\bibitem[Law et~al., 2019]{law2019houseprices}
Law, S., Paige, B., and Russell, C. (2019).
\newblock Take a look around: Using street view and satellite images to
  estimate house prices.
\newblock {\em ACM Transactions on Intelligent Systems and Technology},
  10(5):54:1--54:19.

\bibitem[Li et~al., 2026]{li2026stcl}
Li, Y., Huang, Y., and Zhang, F. (2026).
\newblock Learning street view representations based on a spatiotemporal
  contrastive learning framework.
\newblock {\em Computers, Environment and Urban Systems}, 125:102393.

\bibitem[Li et~al., 2024]{li2024streetviewllm}
Li, Z., Xu, J., Wang, S., Wu, Y., and Li, H. (2024).
\newblock {StreetviewLLM}: Extracting geographic information using a
  chain-of-thought multimodal large language model.
\newblock arXiv preprint arXiv:2411.14476.

\bibitem[Liang et~al., 2025]{liang2025openfacades}
Liang, X., Xie, J., Zhao, T., Stouffs, R., and Biljecki, F. (2025).
\newblock {OpenFACADES}: An open framework for architectural caption and
  attribute data enrichment via street view imagery.
\newblock {\em ISPRS Journal of Photogrammetry and Remote Sensing},
  230:918--942.

\bibitem[Liu et~al., 2026]{liu2026citylens}
Liu, T., Pang, H., Zhang, X., Ouyang, T., Zhang, Z., Feng, J., Li, Y., and Hui,
  P. (2026).
\newblock {CityLens}: Evaluating large vision-language models for urban
  socioeconomic sensing.
\newblock In {\em International Conference on Learning Representations}.

\bibitem[Liu et~al., 2025]{liu2025urbanchange}
Liu, Y., Wang, Z., Ren, S., Chen, R., Shen, Y., and Biljecki, F. (2025).
\newblock Physical urban change and its socio-environmental impact: Insights
  from street view imagery.
\newblock {\em Computers, Environment and Urban Systems}, 119:102284.

\bibitem[Maeda et~al., 2018]{maeda2018roaddamage}
Maeda, H., Sekimoto, Y., Seto, T., Kashiyama, T., and Omata, H. (2018).
\newblock Road damage detection and classification using deep neural networks
  with smartphone images.
\newblock {\em Computer-Aided Civil and Infrastructure Engineering},
  33(12):1127--1141.

\bibitem[Malekzadeh et~al., 2025]{malekzadeh2025attractiveness}
Malekzadeh, M., Willberg, E., Torkko, J., and Toivonen, T. (2025).
\newblock Urban attractiveness according to {ChatGPT}: Contrasting {AI} and
  human insights.
\newblock {\em Computers, Environment and Urban Systems}, 117:102243.

\bibitem[Manvi et~al., 2024]{manvi2024geographic}
Manvi, R., Khanna, S., Burke, M., Lobell, D.~B., and Ermon, S. (2024).
\newblock Large language models are geographically biased.
\newblock In {\em Proceedings of the 41st International Conference on Machine
  Learning}, volume 235 of {\em Proceedings of Machine Learning Research},
  pages 34654--34669. PMLR.

\bibitem[Milojevic-Dupont et~al., 2020]{milojevic2020heights}
Milojevic-Dupont, N., Hans, N., Kaack, L.~H., Zumwald, M., Andrieux, F.,
  de~Barros~Soares, D., Lohrey, S., Pichler, P.-P., and Creutzig, F. (2020).
\newblock Learning from urban form to predict building heights.
\newblock {\em PLOS ONE}, 15(12):e0242010.

\bibitem[Moayeri et~al., 2024]{moayeri2024worldbench}
Moayeri, M., Tabassi, E., and Feizi, S. (2024).
\newblock {WorldBench}: Quantifying geographic disparities in {LLM} factual
  recall.
\newblock In {\em The 2024 ACM Conference on Fairness, Accountability, and
  Transparency}, pages 1211--1228. Association for Computing Machinery.

\bibitem[Ploton et~al., 2020]{ploton2020spatial}
Ploton, P., Mortier, F., R{\'e}jou-M{\'e}chain, M., Barbier, N., Picard, N.,
  Rossi, V., Dormann, C., Cornu, G., Viennois, G., Bayol, N., Lyapustin, A.,
  Gourlet-Fleury, S., and P{\'e}lissier, R. (2020).
\newblock Spatial validation reveals poor predictive performance of large-scale
  ecological mapping models.
\newblock {\em Nature Communications}, 11(1):4540.

\bibitem[Ren et~al., 2024]{ren2024svrdd}
Ren, M., Zhang, X., Zhi, X., Wei, Y., and Feng, Z. (2024).
\newblock An annotated street view image dataset for automated road damage
  detection.
\newblock {\em Scientific Data}, 11(1):407.

\bibitem[Roberts et~al., 2017]{roberts2017crossvalidation}
Roberts, D.~R., Bahn, V., Ciuti, S., Boyce, M.~S., Elith, J., Guillera-Arroita,
  G., Hauenstein, S., Lahoz-Monfort, J.~J., Schr{\"o}der, B., Thuiller, W.,
  Warton, D.~I., Wintle, B.~A., Hartig, F., and Dormann, C.~F. (2017).
\newblock Cross-validation strategies for data with temporal, spatial,
  hierarchical, or phylogenetic structure.
\newblock {\em Ecography}, 40(8):913--929.

\bibitem[Saha et~al., 2019]{saha2019projectsidewalk}
Saha, M., Saugstad, M., Maddali, H.~T., Zeng, A., Holland, R., Bower, S., Dash,
  A., Chen, S., Li, A., Hara, K., and Froehlich, J. (2019).
\newblock {Project Sidewalk}: A web-based crowdsourcing tool for collecting
  sidewalk accessibility data at scale.
\newblock In {\em Proceedings of the 2019 CHI Conference on Human Factors in
  Computing Systems}, pages 1--14. Association for Computing Machinery.

\bibitem[Su et~al., 2026]{su2026buildingsense}
Su, P., Chen, R., Xu, H., Huang, W., Deng, X., Li, S., Yan, W., Wu, H., and
  Liu, C. (2026).
\newblock {BuildingSense}: a new multimodal building function classification
  dataset.
\newblock {\em Earth System Science Data}, 18(4):2609--2634.

\bibitem[Tobler, 1970]{tobler1970growth}
Tobler, W.~R. (1970).
\newblock A computer movie simulating urban growth in the {Detroit} region.
\newblock {\em Economic Geography}, 46(sup1):234--240.

\bibitem[Xie et~al., 2024]{xie2024knowledgeconflicts}
Xie, J., Zhang, K., Chen, J., Lou, R., and Su, Y. (2024).
\newblock Adaptive chameleon or stubborn sloth: Revealing the behavior of large
  language models in knowledge conflicts.
\newblock In {\em International Conference on Learning Representations}.

\bibitem[Yang et~al., 2024]{yang2024commitments}
Yang, Z., Lin, X., He, Q., Huang, Z., Liu, Z., Jiang, H., Shu, P., Wu, Z., Li,
  Y., Law, S., Mai, G., Liu, T., and Yang, T. (2024).
\newblock Examining the commitments and difficulties inherent in multimodal
  foundation models for street view imagery.
\newblock arXiv preprint arXiv:2408.12821.

\bibitem[Zhao et~al., 2025]{zhao2025seasonal}
Zhao, T., Liang, X., Biljecki, F., Tu, W., Cao, J., Li, X., and Yi, S. (2025).
\newblock Quantifying seasonal bias in street view imagery for urban form
  assessment: A global analysis of 40 cities.
\newblock {\em Computers, Environment and Urban Systems}, 120:102302.

\end{thebibliography}
